\documentclass[sigconf]{acmart}
\usepackage{multirow}
\AtBeginDocument{%
  }

\copyrightyear{2025}
\acmYear{2025}
\setcopyright{acmlicensed}
\acmConference[MUWS '25] {Proceedings of the 4th International Workshop on Multimodal Human Understanding for the Web and Social Median}{October 27--31, 2025}{Dublin, Ireland.}
\acmBooktitle{Proceedings of the 4th International Workshop on Multimodal Human Understanding for the Web and Social Media (MUWS '25), October 27--31, 2025, Dublin, Ireland}
\acmISBN{979-8-4007-1838-0/2025/10}
\acmDOI{10.1145/XXXXXX.XXXXXX}

\begin{document}

\title{Revealing Temporal Label Noise in Multimodal Hateful Video Classification}
\renewcommand{\shortauthors}{Yang et al.}
\author{Shuonan Yang}
\authornote{These authors contributed equally to this work.}
\affiliation{%
  \institution{Multimodal Intelligence Lab,}
  \department{Department of Computer Science,}
  \institution{University of Exeter,}
  \city{Exeter}
  \country{United Kingdom}
}
\email{sy446@exeter.ac.uk}

\author{Tailin Chen}
\authornotemark[1] 
\affiliation{%
  \institution{Multimodal Intelligence Lab,}
  \department{Department of Computer Science,}
  \institution{University of Exeter,}
  \city{Exeter}
  \country{United Kingdom}
}
\email{T.Chen2@exeter.ac.uk}

\author{Rahul Singh}
\authornotemark[1] 
\affiliation{%
  \institution{Multimodal Intelligence Lab,}
  \department{Department of Computer Science,}
  \institution{University of Exeter,}
  \city{Exeter}
  \country{United Kingdom}
}
\email{R.Singh3@exeter.ac.uk}

\author{Jiangbei Yue}
\affiliation{%
  \institution{Multimodal Intelligence Lab,}
  \department{Department of Computer Science,}
  \institution{University of Exeter,}
  \city{Exeter}
  \country{United Kingdom}
}
\email{J.Yue@exeter.ac.uk}

\author{Jianbo Jiao}
\affiliation{%
  \department{Department of Computer Science,}
  \institution{University of Birmingham,}
  \city{Birmingham}
  \country{United Kingdom}
}
\email{j.jiao@bham.ac.uk}

\author{Zeyu Fu}
\authornote{Corresponding author.}
\affiliation{%
  \institution{Multimodal Intelligence Lab,}
  \department{Department of Computer Science,}
  \institution{University of Exeter,}
  \city{Exeter}
  \country{United Kingdom}
}
\email{Z.Fu@exeter.ac.uk}

\begin{abstract} 
  The rapid proliferation of online multimedia content has intensified the spread of hate speech, presenting critical societal and regulatory challenges. While recent work has advanced multimodal hateful video detection, most approaches rely on coarse, video-level annotations that overlook the temporal granularity of hateful content. This introduces substantial label noise, as videos annotated as hateful often contain long non-hateful segments. In this paper, we investigate the impact of such label ambiguity through a fine-grained approach. Specifically, we trim hateful videos from the HateMM and MultiHateClip English datasets using annotated timestamps to isolate explicitly hateful segments. We then conduct an exploratory analysis of these trimmed segments to examine the distribution and characteristics of both hateful and non-hateful content. This analysis highlights the degree of semantic overlap and the confusion introduced by coarse, video-level annotations. Finally, controlled experiments demonstrated that time-stamp noise fundamentally alters model decision boundaries and weakens classification confidence, highlighting the inherent context dependency and temporal continuity of hate speech expression. Our findings provide new insights into the temporal dynamics of multimodal hateful videos and highlight the need for temporally aware models and benchmarks for improved robustness and interpretability. Code and data are available at https://github.com/Multimodal-Intelligence-Lab-MIL/HatefulVideoLabelNoise.
 
    {\color{red}\textbf{Warning}: This research work contains sensitive materials that many will find disturbing and offensive. Regrettably, such material is unavoidable due to the nature of the study.}
\end{abstract}





\maketitle

\section{Introduction}
The proliferation of online hate speech poses a serious threat to digital security and community cohesion, prompting researchers to explore detection and content moderation methods. Early studies focused on text content, employing manually designed feature extraction methods, which were later enhanced with deep learning techniques to effectively identify hate speech patterns in linguistic contexts \cite{Chen_2012, Davidson_2017, RNN_2020, gamback-sikdar-2017-using, zimmerman-etal-2018-improving}. The evolution toward multimedia-rich online platforms has shifted hateful content increasingly toward multimodal formats, particularly memes and videos. Meme-focused detection frameworks, such as MOMENTA \cite{MOMENTA}, MemeFier \cite{MemeFier}, and prompt-based models \cite{Prompt_based}, have successfully leveraged combined visual-textual analysis. However, video-based hate speech detection introduced greater complexity. Das et al. \cite{das2023hatemm} developed the first hateful video dataset HateMM, while Zhang et al. \cite{CMFusion} employed complex cross-attention fusion techniques. Wang et al. \cite{Wang2025} further improved performance through integrating memes and fine-tuning vision-language models.

Existing video-based hate detection methods primarily follow two paradigms. The first is the traditional paradigm proposed by HateMM \cite{das2023hatemm}, which equips each modality with a dedicated encoder and then integrates them through attention mechanisms or concatenation fusion. The second is an emerging large language models (LLM) based paradigm that adopts LLM \cite{Wang2025}, but this direction is still in its exploratory stage in the field of hate detection. 
Both paradigms primarily rely on coarse-grained video-level annotations, labelling the entire video as either hateful or non-hateful. These methods ignore the inherent temporal granularity of hate speech. According to our analysis, as shown in Table~\ref{tab:segment_stats} and Figure~\ref{fig:seg-lengths}, in HateMM \cite{das2023hatemm}, hateful videos average 2.56 minutes in length, yet hateful segments span only 1.71 minutes on average, indicating 33\% non-hateful content within videos labelled as hate. Similarly, MultiHateClip-English \cite{wang2024multihateclip} exhibits temporal discrepancies, with trimmed hate segments averaging 24.50 seconds compared to 9.36 seconds for non-hate segments extracted from the same videos. This phenomenon indicates that hate videos contain a large amount of irrelevant content, and the key issue of label noise caused by mismatches in time granularity in video-level labels deserves further exploration and verification in the analysis of model detection performance and robustness.

\begin{figure*}[t]
  \centering
  \includegraphics[width=0.8\linewidth]{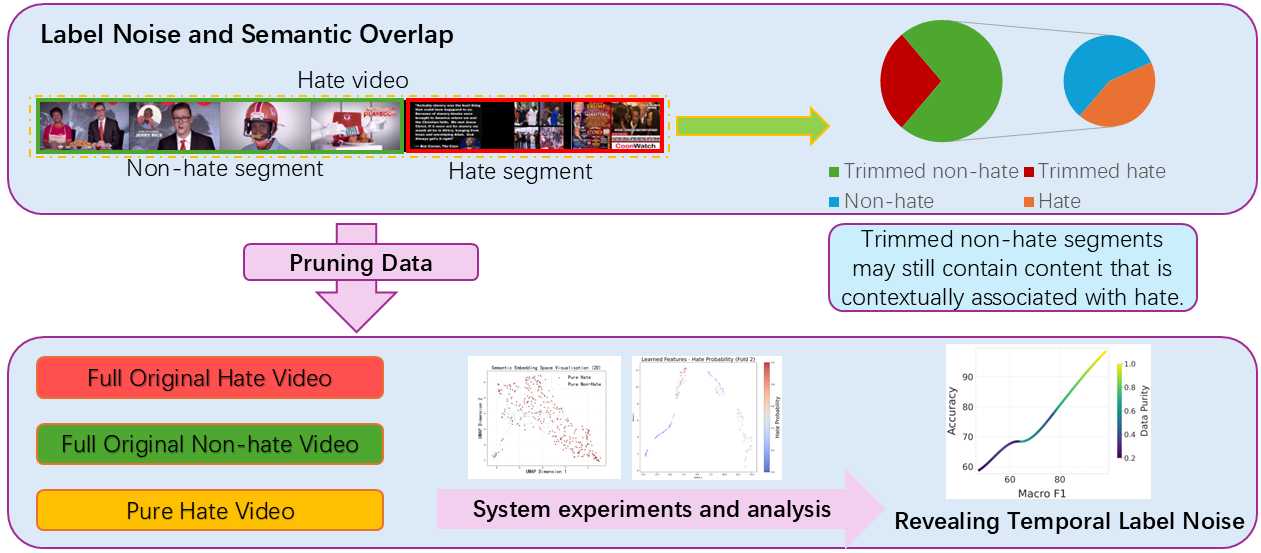}
  \caption{Our overall research pipeline. We first filter out hateful segments from hate videos through time annotation and perform quantitative analysis. Subsequently, the organised data subset is used for experiments to reveal how label noise affects model performanc.}
  \label{fig:pipline}
\end{figure*}

To address these questions, we conducted the first systematic study on temporal label noise in multimodal hate video classification. As shown in Figure~\ref{fig:pipline}, we leverage annotated timestamps from HateMM and MultiHateClip-English datasets to extract trimmed hate and trimmed non-hate segments, enabling controlled analysis of semantic boundaries and detection performance. Through comprehensive lexical analysis, semantic embedding visualisation, and controlled experimental evaluation, we quantify the impact of temporal label noise on classification accuracy and model robustness.
Our key contributions include: First, through empirical quantification and semantic analysis, we reveal that label noise represents systematic rather than random contamination, and semantic overlap leads to classification challenges; Second, through controlled experiments, we demonstrate that methods based on clean segment-level detection outperform noise video-level methods, achieving macro F1 score improvements of 19.34\% and 30.45\%, respectively. These findings highlight the urgency of developing time-aware detection models and improved annotation strategies to address the localised characteristics of hate speech expression.

\section{Related Work}
\subsection{Hate Speech Detection}
Hate speech detection technology has gradually evolved from early text-based methods to complex multimodal frameworks \cite{Chen_2012, Davidson_2017, das2023hatemm}. Traditional methods primarily rely on linguistic features and semantic analysis of text data, employing techniques ranging from dictionary-based models to deep learning architectures such as recurrent neural networks \cite{RNN_2020}. However, with the widespread adoption of multimedia content, text-based detection systems have revealed fundamental limitations, particularly their inability to capture visual and auditory cues. 

Recent studies have demonstrated that multimodal approaches offer significant advantages in identifying hateful content across multiple media formats. Systems such as MOMENTA \cite{MOMENTA} and MemeFier \cite{MemeFier} have made significant progress by focusing on visual forms of hate speech that combine subtle textual and visual elements (especially memes). These models effectively align the semantic representations of different modalities through cross-modal attention mechanisms, achieving significant performance improvements over single-modal baselines. Recently, research has expanded from static images to dynamic content. Das et al. \cite{das2023hatemm} proposed HateMM, the first large-scale multimodal dataset specifically for hateful video content, marking an important milestone in the field of video-based hate speech detection. To address the inherent challenges of heterogeneous modality fusion, Zhang et al. \cite{CMFusion} proposed an advanced cross-modal fusion mechanism based on a complex attention architecture, which effectively models complex cross-modal interactions while maintaining computational efficiency. Additionally, Wang et al. \cite{Wang2025} improved detection performance by combining emoji understanding with fine-tuned visual-language models, while enhancing cross-modal alignment and generalisation capabilities. In a multilingual setting, the same research team proposed the MultiHateClip \cite{wang2024multihateclip} dataset, which covers Chinese and English hate speech videos, thereby extending the applicability of hate speech detection systems to non-English media environments. Lang et al. \cite{MoRE} introduces a retrieval-augmented mixture-of-experts framework for short video hate detection, leveraging modality-specific experts and dynamic fusion to address the evolving and multimodal nature of hateful content.
These advances highlight the critical role of modality-aware design and advanced fusion strategies in addressing the evolving characteristics of online media. 

However, existing methods mainly focus on optimising model architecture and fusion mechanisms, which essentially rely on video-level labels and treat the entire video as a unified entity of hateful or non-hateful content, ignoring the noise impact caused by the inherent temporal granularity of hateful content expression.

\begin{table*}[t]
\caption{Descriptive statistics of trimmed datasets showing segment distribution and temporal characteristics}
  \label{tab:segment_stats}
  \centering
  \begin{tabular}{@{}llrrrr@{}}
    \toprule
    \textbf{Dataset} & \textbf{Segment Type} & \textbf{Count} & \textbf{Percentage (\%)} & \textbf{Total Duration} & \textbf{Avg Length (sec)} \\
    \midrule
    HateMM         & Trimmed hate     & 790   & 41.36 & 12.45 hours& 57.24 \\
    HateMM         & Trimmed non-hate & 1,120 & 58.64 & 6.01 hours  & 19.57 \\
    MHC-English    & Trimmed hate     & 332   & 64.84 & 135.56 min  & 24.50 \\
    MHC-English    & Trimmed non-hate & 180   & 35.16 & 28.07 min   & 9.36  \\
    \bottomrule
  \end{tabular}
\end{table*}
\subsection{Weak Supervision and Label Noise in Video Classification}
The challenge of label noise in video classification has been recognised as a fundamental limitation affecting model performance and generalisation. Label noise arises from various sources, including inconsistent human annotations, imprecise temporal boundaries, and the inherent difficulty of categorising complex video content. Leung et al. \cite{leung_hangling_lable_noise} demonstrated that weakly-labelled data from internet sources can contain substantial noise levels, traditional boosting approaches like AdaBoost suffer significant performance degradation under high noise conditions, particularly when noise levels exceed 20\%. To address these challenges, Multiple Instance Learning (MIL) has emerged as a promising framework for handling label ambiguity in video data \cite{MIL_label_noise_1, leung_hangling_lable_noise}. In MIL formulations, positive training examples are grouped into bags, where each bag is considered positive if it contains at least one true positive instance. This approach naturally accommodates the temporal structure of videos, where hateful content may be localised within specific segments whilst the entire video receives a coarse label. The Noisy OR model employed in MILBoost \cite{MILBoost} effectively reduces label noise by leveraging the assumption that accurately labelled instances within each bag can compensate for mislabelled examples.

Although these methods have demonstrated effectiveness in general video classification tasks, hateful video content faces unique challenges that are fundamentally different from traditional label noise scenarios. Hate speech in videos is highly context-dependent and often requires temporal consistency across multiple frames to establish semantic meaning \cite{CMFusion, MoRE}. Aggressive gestures or inflammatory text displayed in a single frame may only be interpreted as hate speech within a broader temporal context (spanning dozens of adjacent frames). While temporal granularity issues are prevalent in hate speech detection, no previous studies have specifically investigated the impact of such context-dependent label noise on video classification performance.

\section{Dataset Preprocessing and Trimming}
The literature review reveals that while multimodal hate detection has advanced significantly, the impact of temporal label granularity remains unexplored. To address this gap, we conduct a systematic investigation using established datasets with temporal annotations, enabling controlled analysis of label noise effects on model performance. This study uses two benchmark multimodal hateful video datasets: HateMM \cite{das2023hatemm} and MultiHateClip-English \cite{wang2024multihateclip}. Both offer video-level labels and temporal annotations, enabling fine-grained analysis of hateful segments.
\begin{figure}[t]
  \centering
  \includegraphics[width=0.9\linewidth]{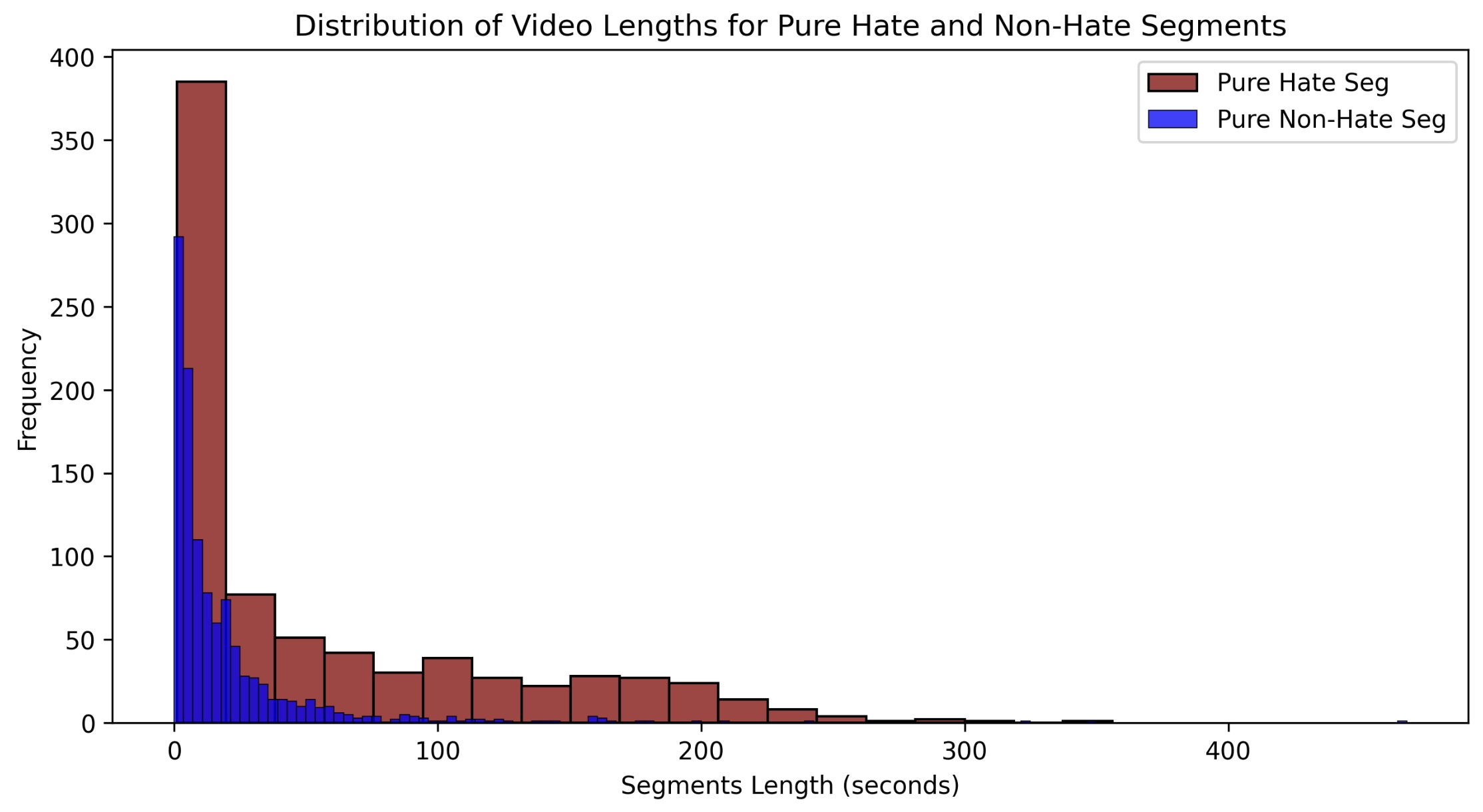}
  \includegraphics[width=0.9\linewidth]{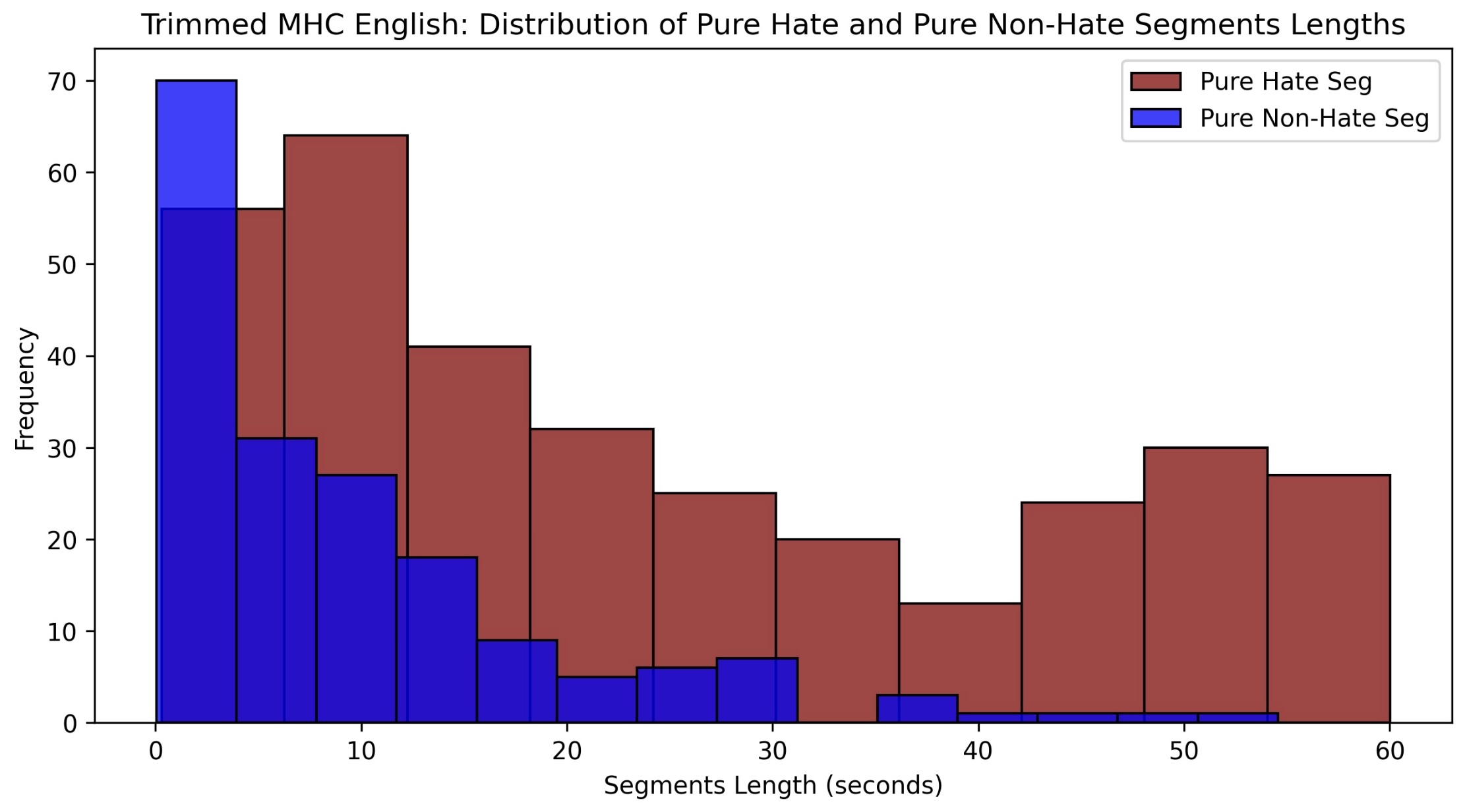}
  \caption{Distribution of segment lengths in the trimmed HateMM dataset. 
    }
  \label{fig:seg-lengths}
\end{figure}

For the MultiHateClip dataset, we follow the binary classification setting established in MultiHateClip\cite{wang2024multihateclip}, where \textit{hateful} and \textit{offensive} labels are merged into a single \textit{hateful} class, resulting in a binary label space. We apply a trimmed segmentation strategy to all videos labeled as hateful in both the HateMM and MultiHateClip-English datasets. Specifically, for each annotated hateful span represented by a timestamp pair $(s_i, e_i)$, where $s_i$ and $e_i$ denote the start and end times of the span, we extract two distinct types of segments:
\begin{itemize}
    \item \text{Trimmed hate segments}: The interval $[s_i, e_i]$, containing content explicitly marked as hateful.
    \item \text{Trimmed non-hate segments}: Content occurring before $s_i$ or after $e_i$ within the same hateful video, ensuring no temporal overlap with any annotated hate span.
\end{itemize}

For videos originally labeled as non-hateful, we retain them as complete videos without any temporal segmentation, as these videos contain no annotated hateful spans requiring extraction. This preprocessing strategy enables controlled comparison between: (1) trimmed hate segments extracted from hateful videos, (2) trimmed non-hate segments extracted from the same hateful videos, and (3) complete non-hateful videos. This three-way categorisation reduces label noise inherent in coarse video-level annotations and supports fine-grained analysis of temporal label contamination effects.

As shown in Table~\ref{tab:segment_stats}, the trimmed datasets reveal significant distributional differences between trimmed hate and non-hate segments. HateMM exhibits longer average durations for hate segments (57.24 seconds) compared to non-hate segments (19.57 seconds), whilst MultiHateClip-English shows a similar pattern with hate segments averaging 24.50 seconds versus 9.36 seconds for non-hate content. As illustrated in Figure~\ref{fig:seg-lengths}, the distribution of segment lengths demonstrates that hate segments tend to cluster around longer durations, with HateMM showing a pronounced concentration of hate segments in the 50-100 second range, whilst non-hate segments predominantly occur within shorter temporal windows below 50 seconds. MultiHateClip-English exhibits a similar but more compressed pattern, with hate segments distributed across 10-40 seconds and non-hate segments concentrated in the 5-15 second range. This disparity suggests that hateful content tends to be more sustained and concentrated within specific temporal windows, requiring extended context for full semantic expression.

\section{Data Exploratory Analysis}
\subsection{Lexical Semantic Analysis}
To examine semantic distinctions between trimmed hate and non-hate segments, we conducted lexical analysis using Empath \cite{Fast_2016} categories. We selected 70 categories, excluding irrelevant topics such as technology and entertainment, and reported the top 15 with the most significant differences.
\begin{figure}[t]
  \centering
  \includegraphics[width=\linewidth]{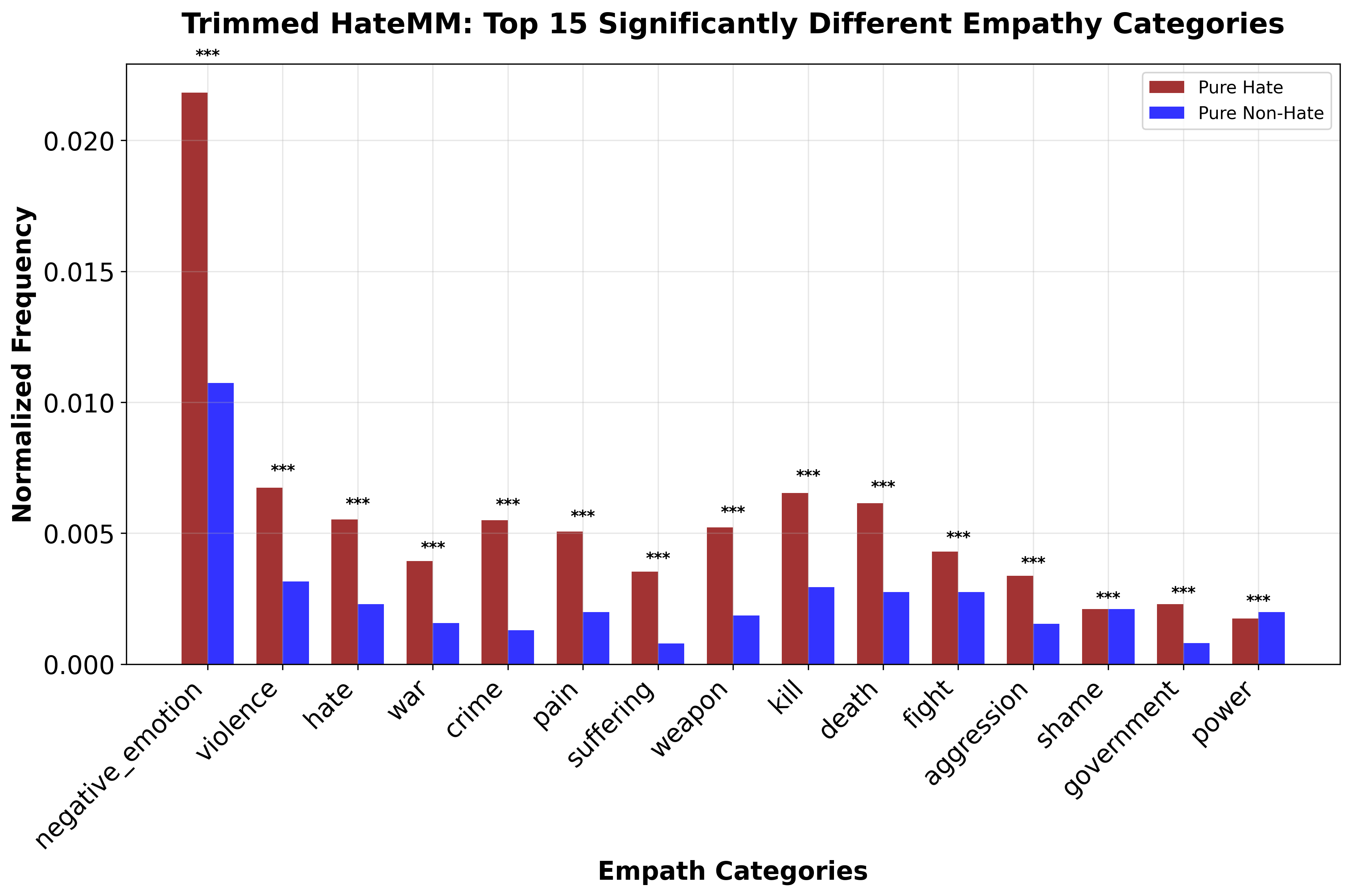}
  \includegraphics[width=\linewidth]{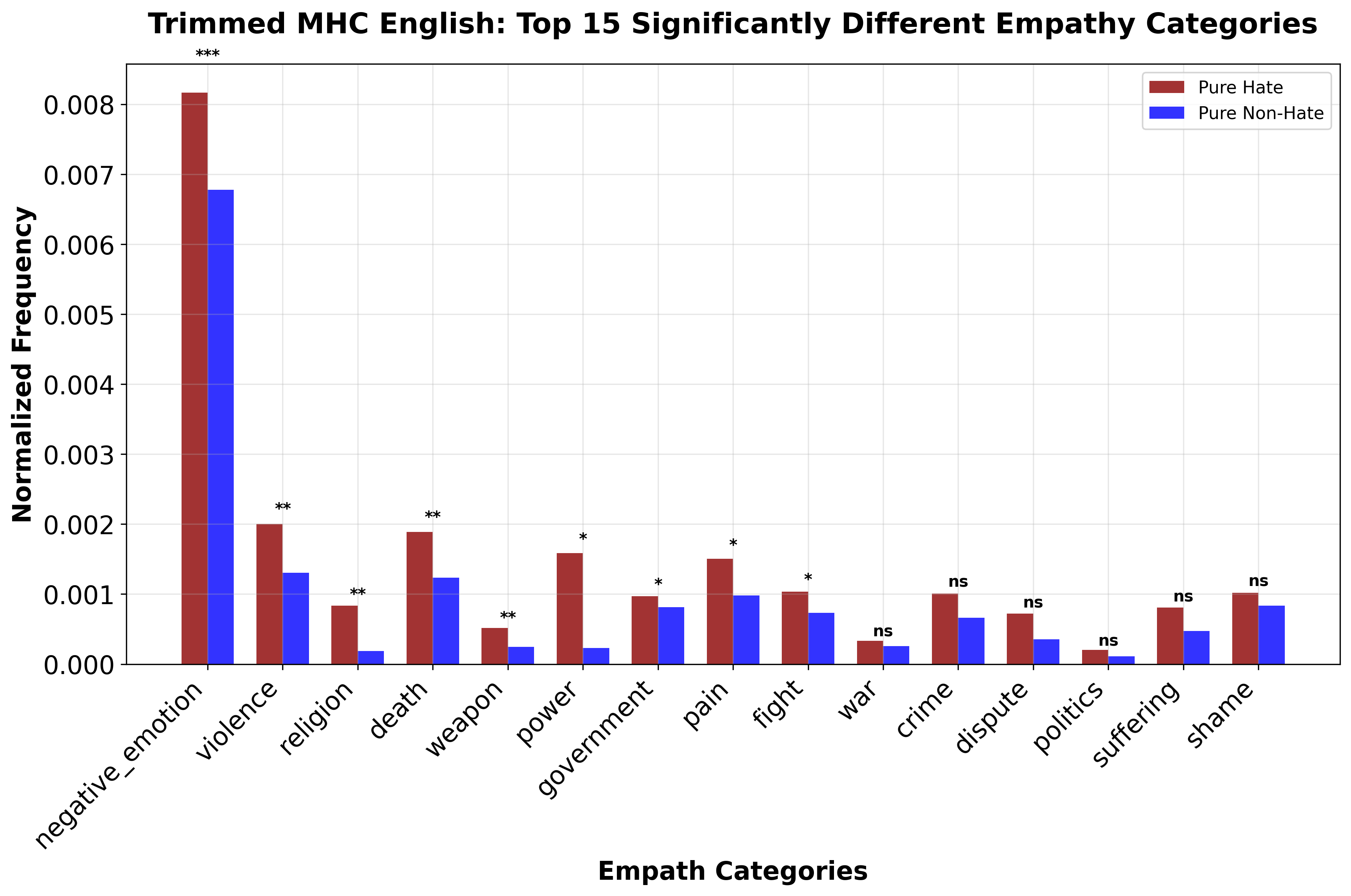}
  \caption{Perform lexical analysis on transcribed text segments classified as trimmed hate and trimmed non-hate using the Empath category. 
  }
  \label{fig:lexical}
\end{figure}
\begin{figure}[t]
  \centering
  \includegraphics[width=0.9\linewidth]{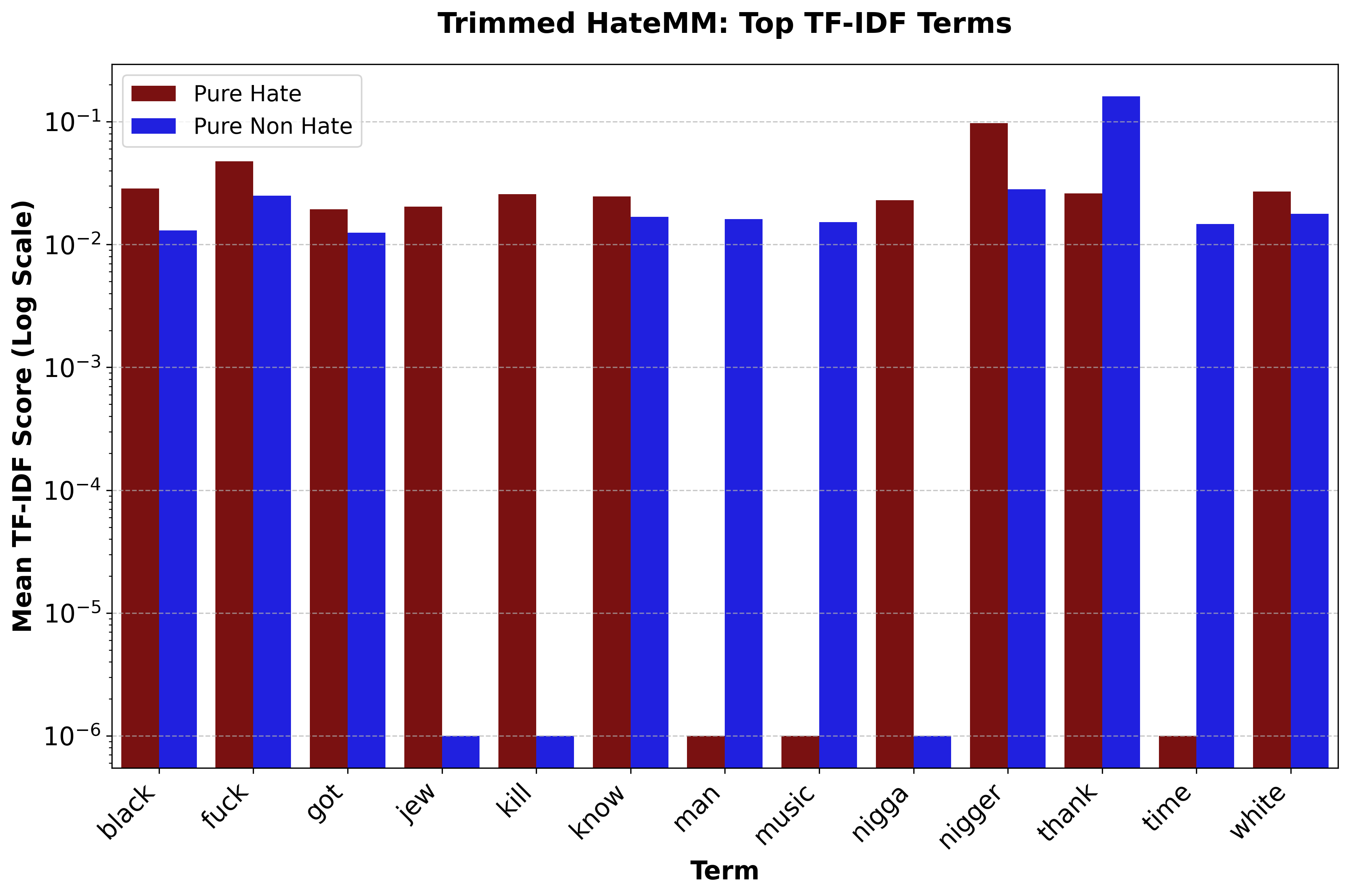}
  \includegraphics[width=0.9\linewidth]{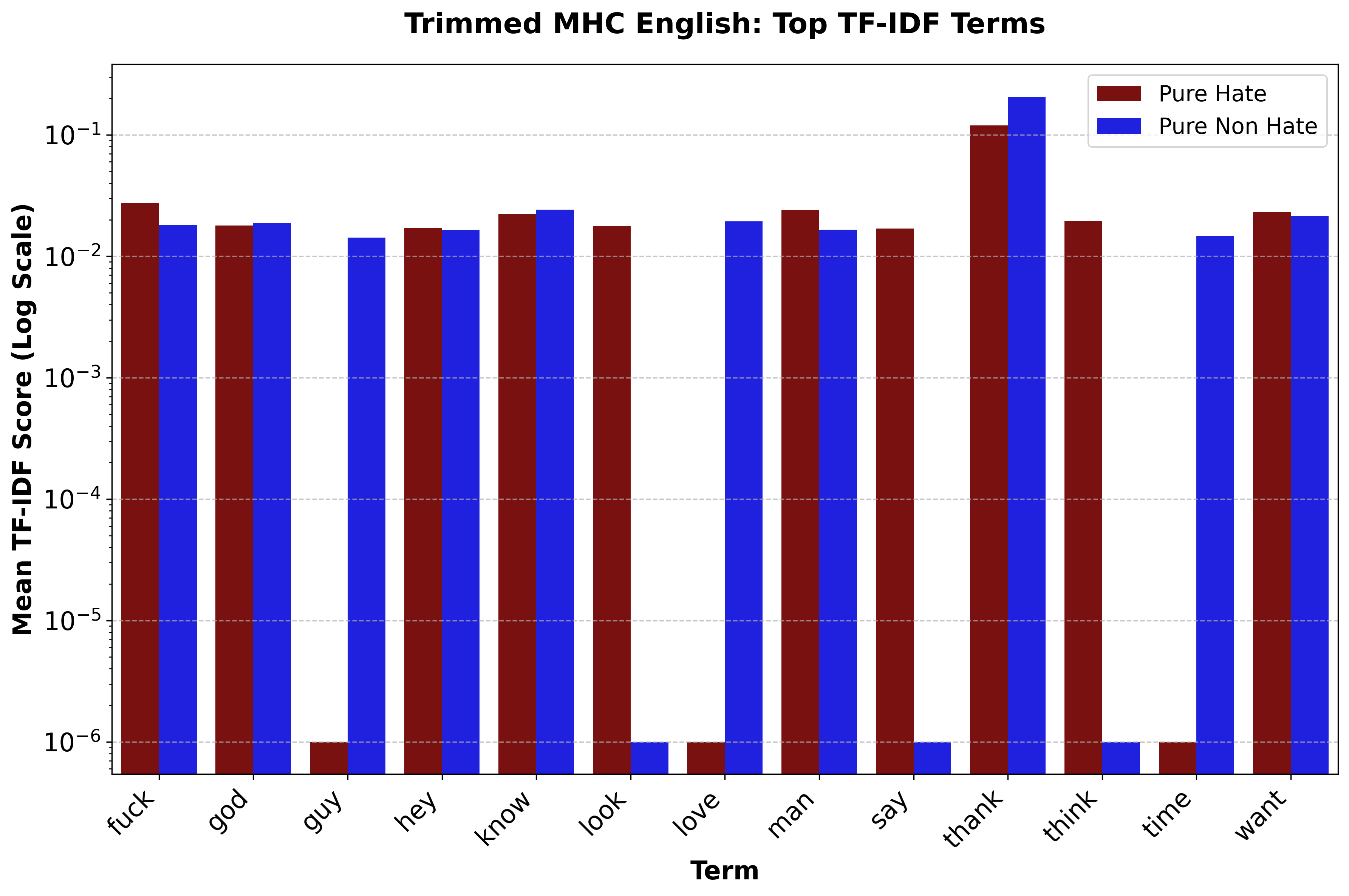}
  \caption{Top TF-IDF terms from trimmed hate and trimmed non-hate segment transcripts. 
}
  \label{fig:TF-IDF}
\end{figure}
\begin{figure}[t]
  \centering
  \includegraphics[width=\linewidth]{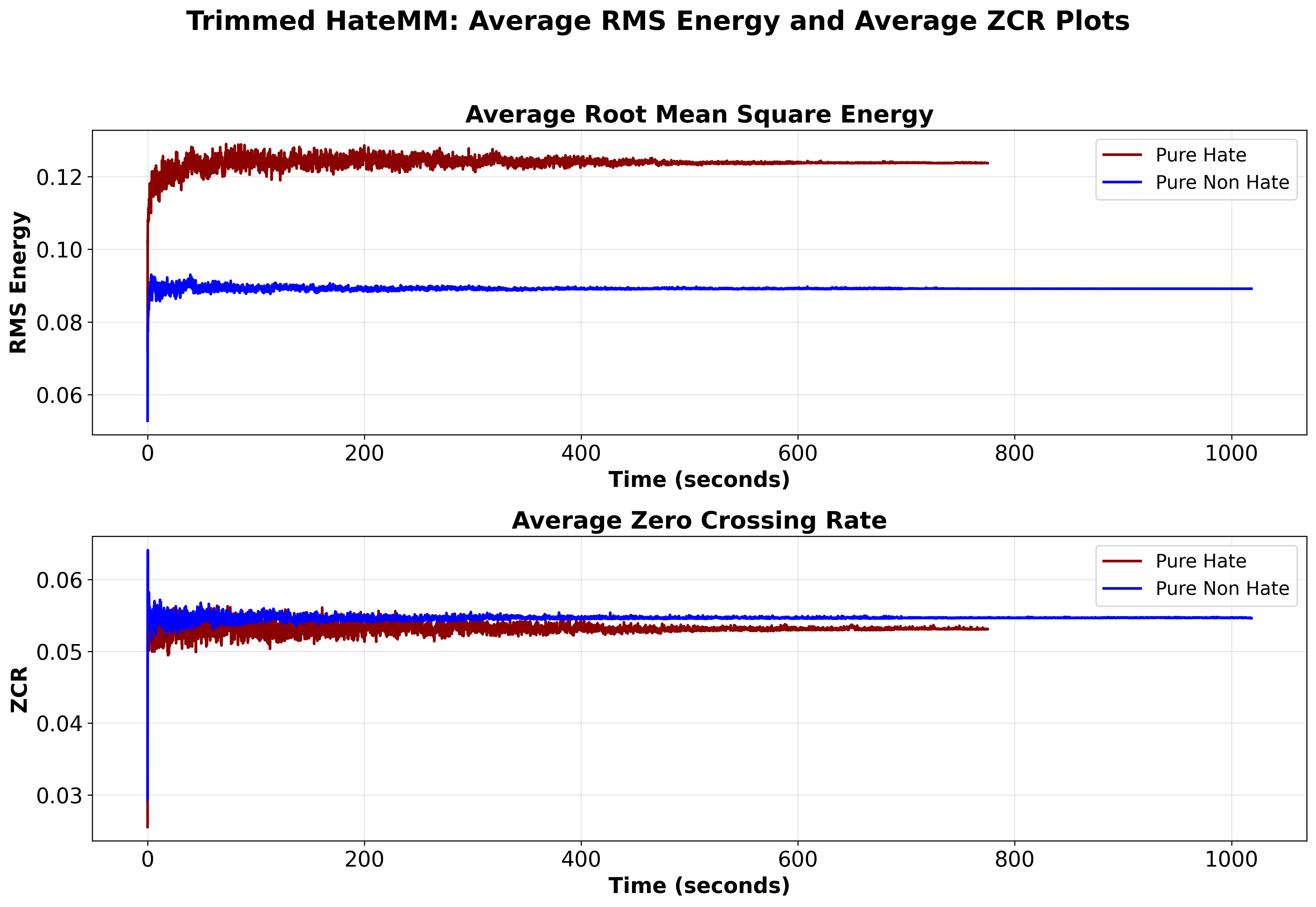}
  \includegraphics[width=\linewidth]{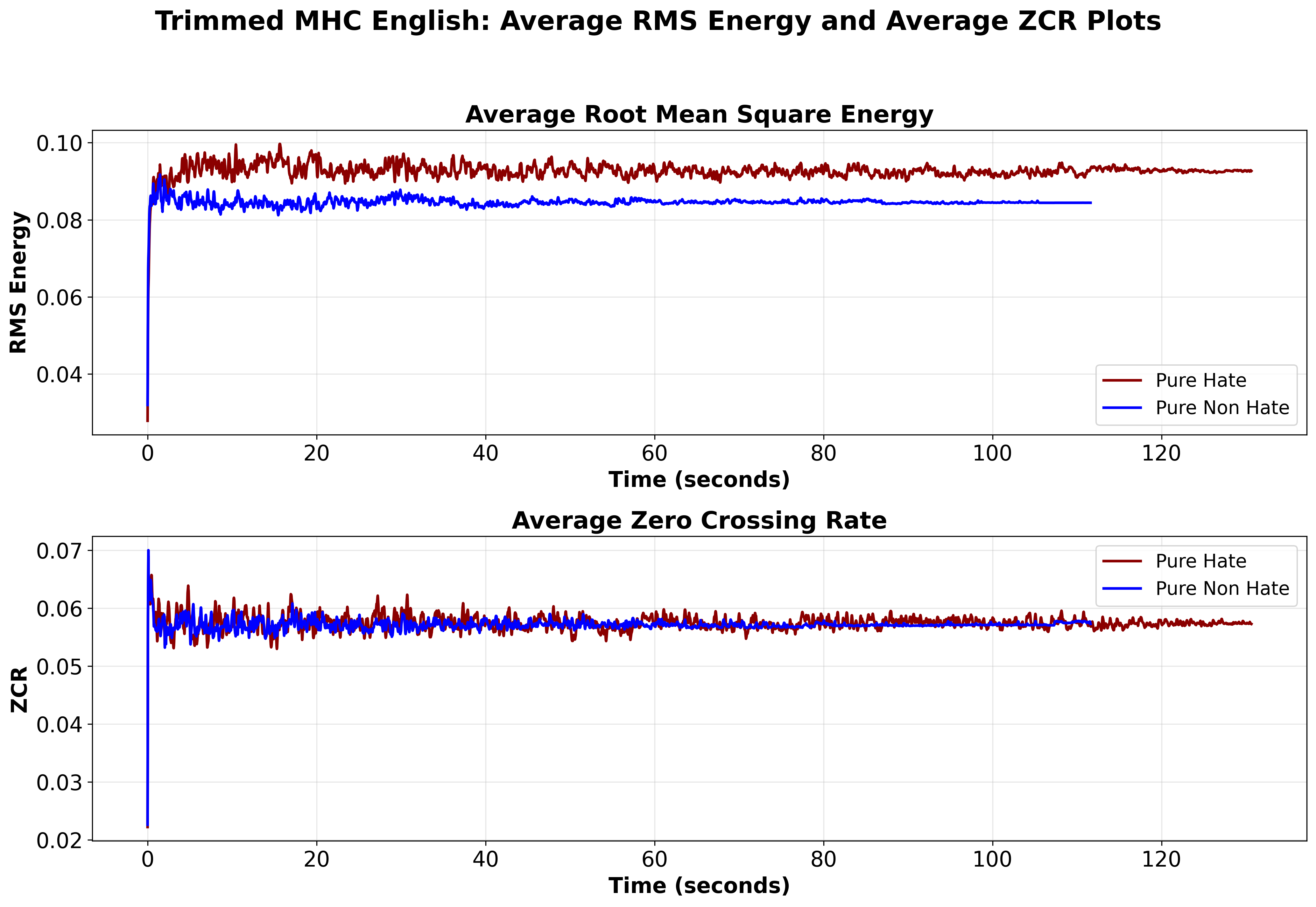}
  \caption{Average Root Mean Square Energy and Zero Crossing Rate Plots}
  \label{fig:avg_rms_zcr}
\end{figure}
\begin{figure*}[t]

  \begin{minipage}[t]{0.24\linewidth}
    \includegraphics[width=\linewidth]{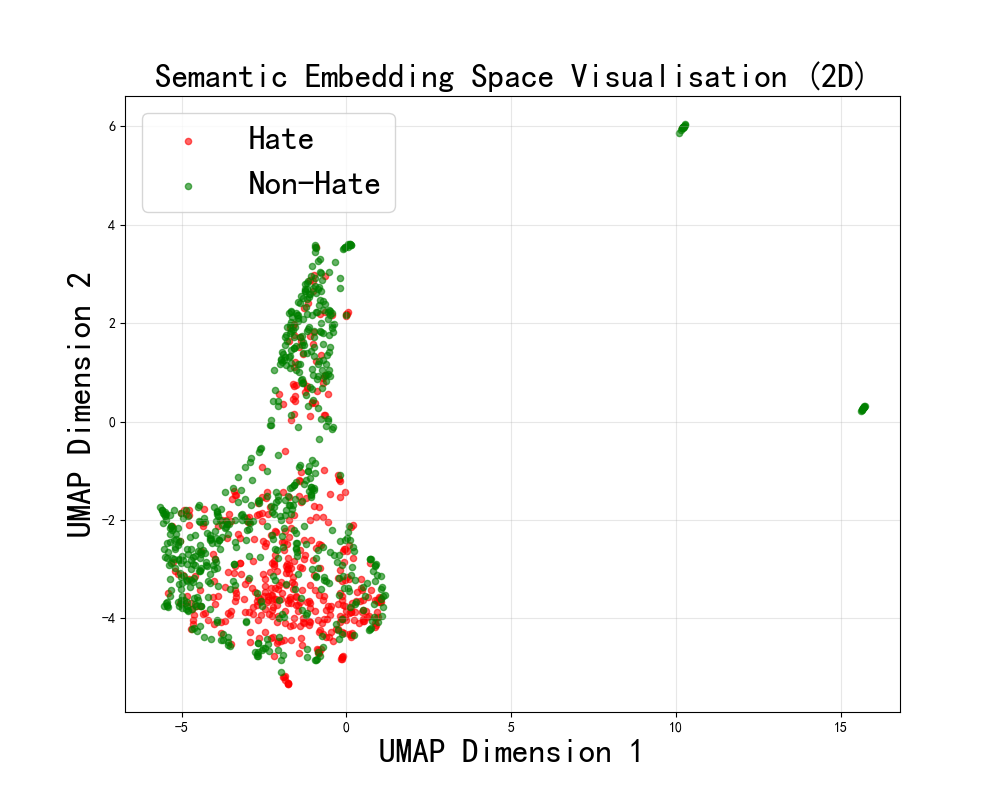}
    \centerline{\small (a) Hate vs. Non-Hate}
  \end{minipage}
  \hfill
  \begin{minipage}[t]{0.24\linewidth}
    \includegraphics[width=\linewidth]{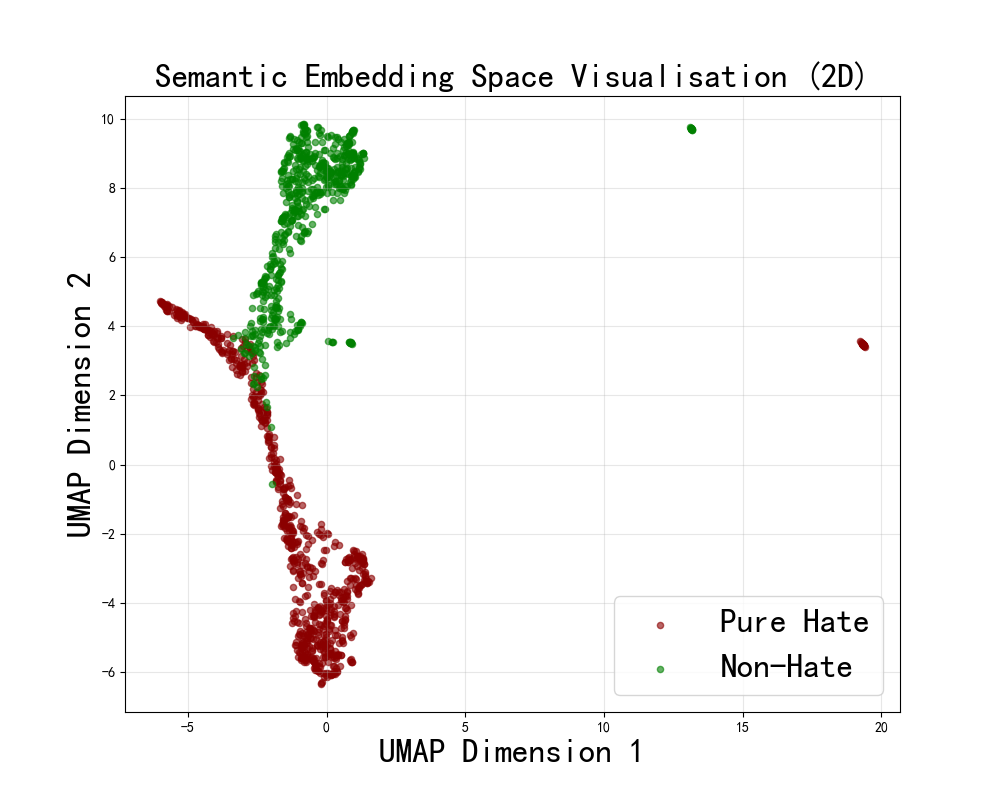}
    \centerline{\small (b) Trimmed Hate vs. Non-Hate}
  \end{minipage}
  \hfill
  \begin{minipage}[t]{0.24\linewidth}
    \includegraphics[width=\linewidth]{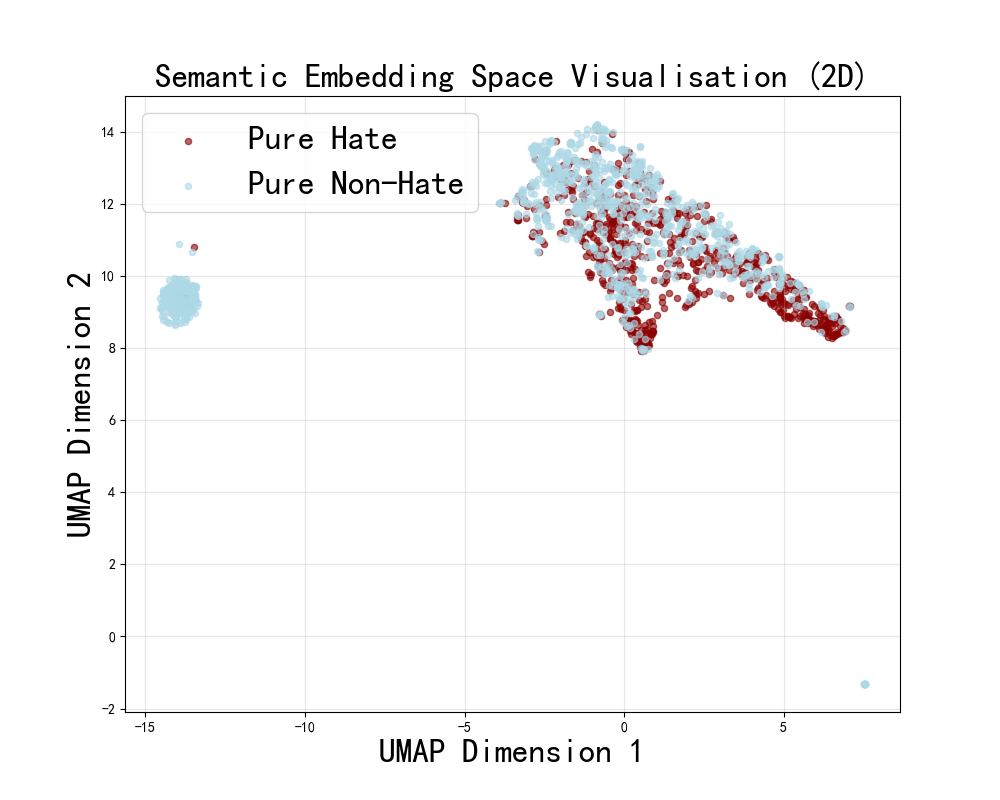}
    \centerline{\small (c) Trimmed Hate vs. Trimmed Non-Hate}
  \end{minipage}
  \hfill
  \begin{minipage}[t]{0.24\linewidth}
    \includegraphics[width=\linewidth]{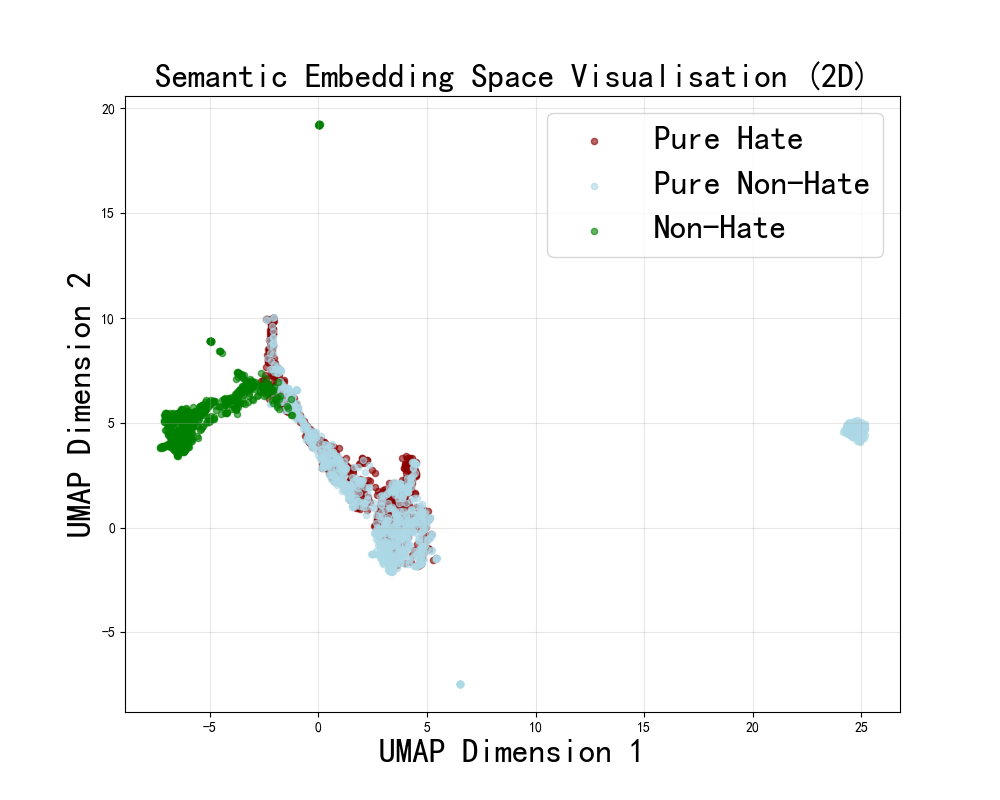}
    \centerline{\small (d) All Three Classes}
  \end{minipage}
  

  \begin{minipage}[t]{0.24\linewidth}
    \includegraphics[width=\linewidth]{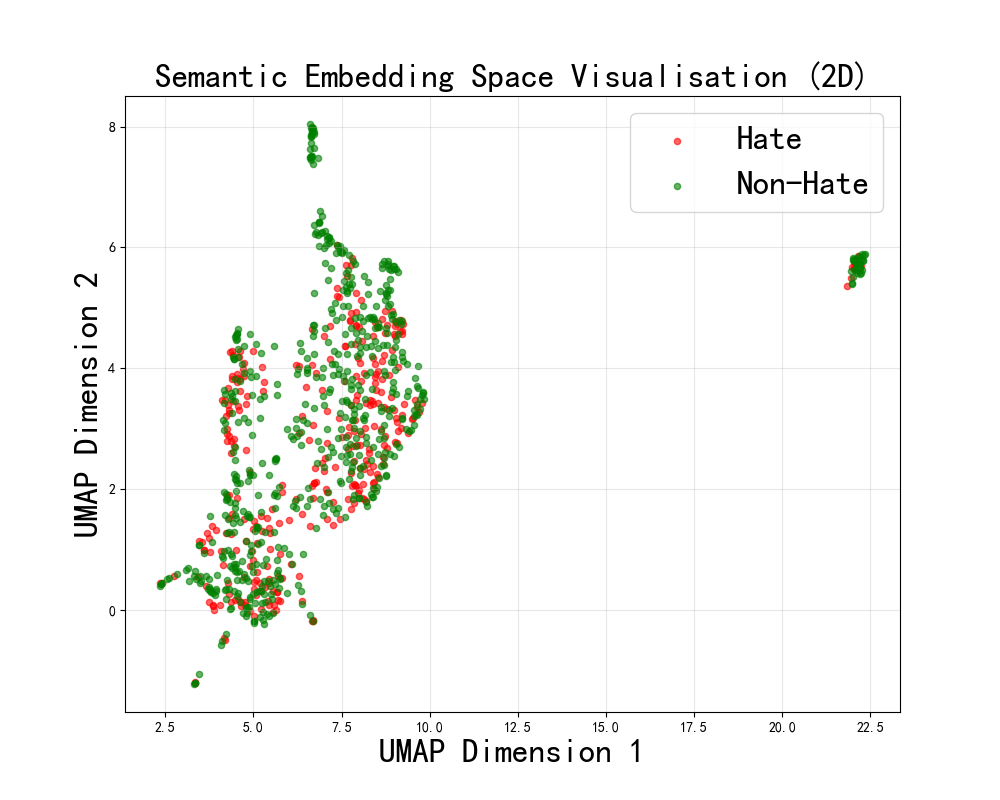}
    \centerline{\small (e) Hate vs. Non-Hate}
  \end{minipage}
  \hfill
  \begin{minipage}[t]{0.24\linewidth}
    \includegraphics[width=\linewidth]{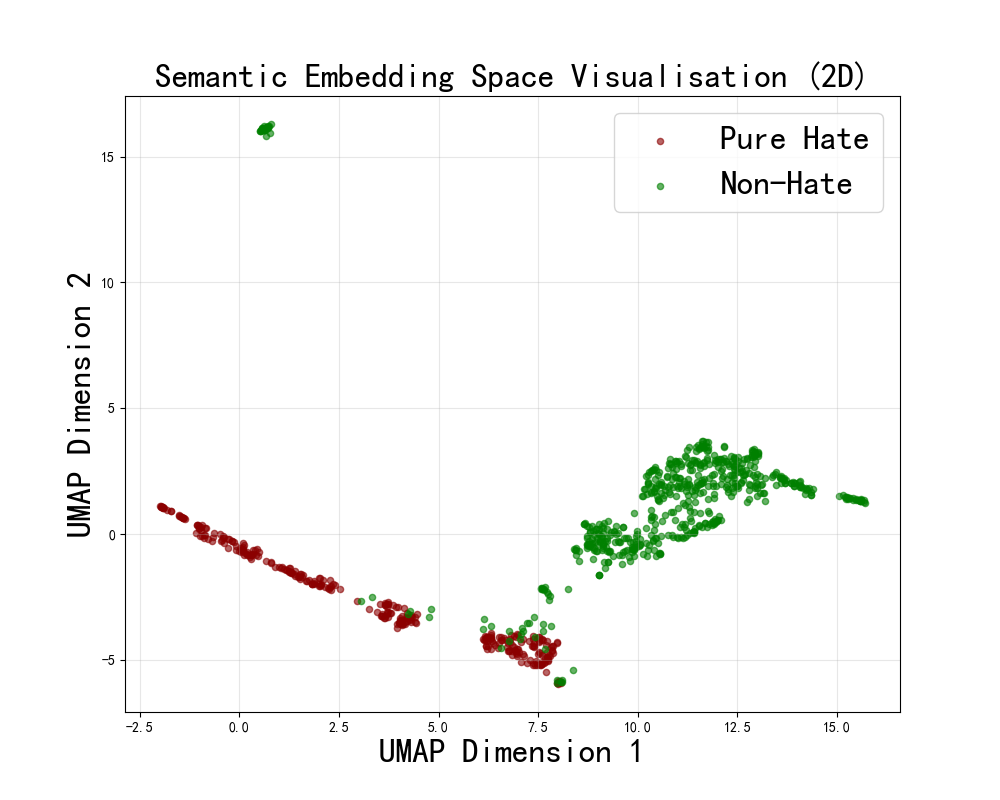}
    \centerline{\small (f) Trimmed Hate vs. Non-Hate}
  \end{minipage}
  \hfill
  \begin{minipage}[t]{0.24\linewidth}
    \includegraphics[width=\linewidth]{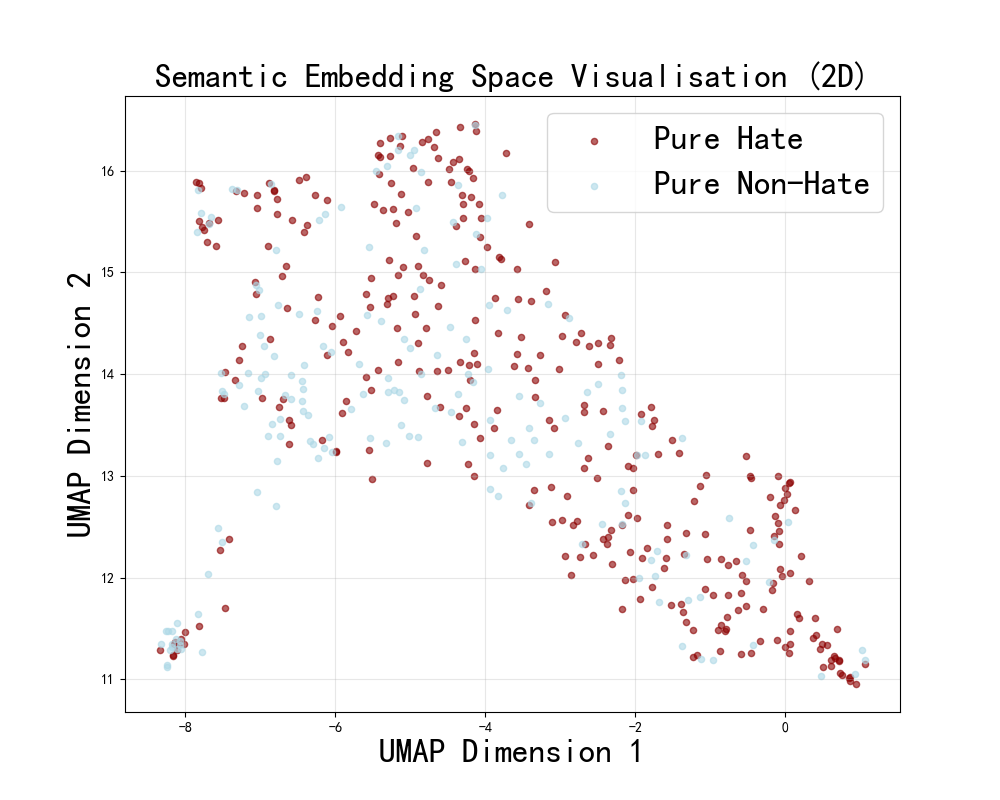}
    \centerline{\small (g) Trimmed Hate vs. Trimmed Non-Hate}
  \end{minipage}
  \hfill
  \begin{minipage}[t]{0.24\linewidth}
    \includegraphics[width=\linewidth]{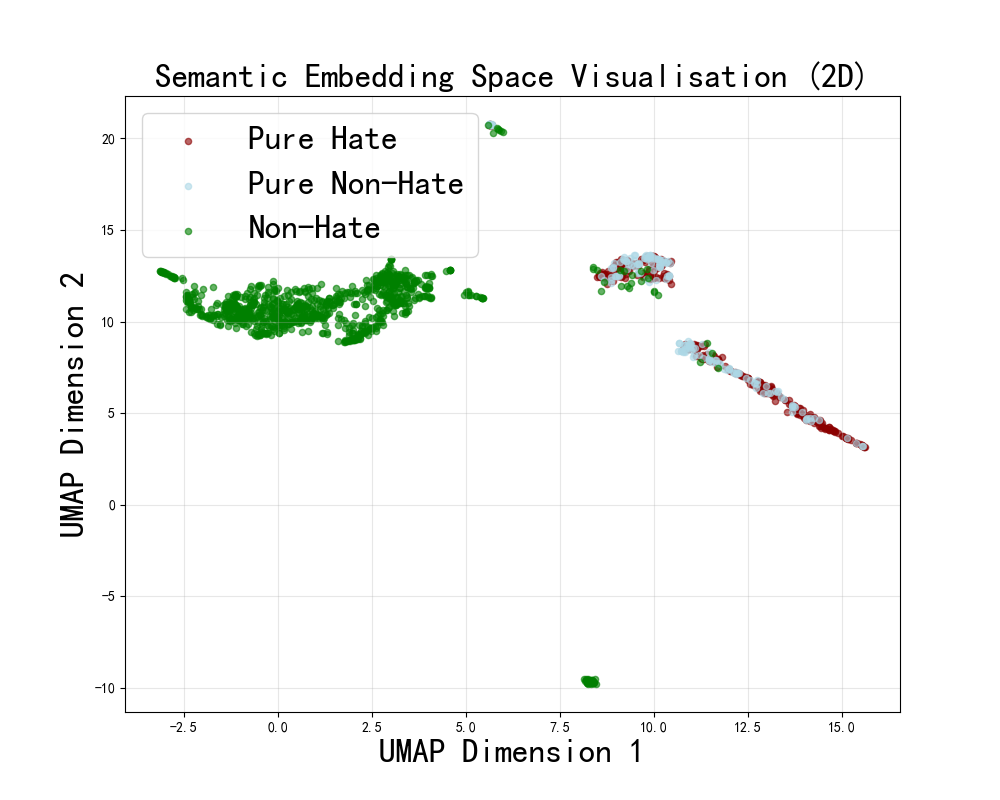}
    \centerline{\small (h) All Three Classes}
  \end{minipage}

  \caption{
    Semantic embedding space visualisations using UMAP for two datasets. Each point represents a transcript segment.
    Top row (a–d): HateMM dataset.
    Bottom row (e–h): MultiHateClip dataset.
    Trimmed Hate: dark red, Trimmed Non-Hate: light blue, Hate: red, Non-Hate: green.
  }
  \label{fig:umap_semantic_space}
\end{figure*}
As illustrated in Figure~\ref{fig:lexical}, trimmed hate segments demonstrate significantly elevated scores across multiple emotionally charged categories compared to non-hate segments (Mann-Whitney U test, $p < 0.05$). HateMM, sourced from BitChute, exhibits pronounced differences with hate segments consistently scoring higher across categories such as 'negative\_emotion', 'violence', 'hate', 'war', 'crime', 'death', 'kill', and 'fight'. MultiHateClip-English, sourced from YouTube, displays a similar directional pattern but with more moderate variations, particularly in 'negative\_emotion', 'violence', 'religion', and 'death' categories. This suggests that BitChute content exhibits more extreme linguistic characteristics compared to YouTube's more moderated environment, reflecting the platforms' differing content policies and user demographics.

Term Frequency-Inverse Document Frequency (TF-IDF) analysis reveals distinct vocabularies between segment types across both platforms, as shown in Figure~\ref{fig:TF-IDF}. HateMM demonstrates stark lexical polarisation, with hate segments dominated by explicit slurs and profanity, whilst non-hate segments feature more conversational language. MultiHateClip-English exhibits more balanced term distributions, though hate segments still show elevated scores for offensive terminology. The platform-specific differences reflect varying content moderation approaches, with YouTube's stricter guidelines potentially constraining the presence of explicitly hateful language compared to BitChute's more permissive environment. However, the presence of certain offensive terms in both segment types across both platforms indicates semantic overlap that may contribute to classification challenges, suggesting that lexical boundaries between hateful and non-hateful content are not always distinct and may exhibit contextual dependencies.

\subsection{Audio Analysis}
We examine acoustic characteristics using Average Root Mean Square (RMS) Energy and Zero Crossing Rate (ZCR) to complement our textual analysis. RMS Energy measures signal intensity, whilst ZCR captures spectral characteristics related to speech noisiness and pitch variation.

\begin{table}[t]
  \centering
  \caption{Person Detection Frequency (\%) in Trimmed Hate (TH) vs Trimmed Non-Hate Segments (TN)}
  \label{tab:person_detection}
  \begin{tabular}{lccc}
    \toprule
    \textbf{Dataset} & \textbf{TH} & \textbf{TN} & \textbf{Difference} \\
    \midrule
    HateMM & 65.81 & 65.23 & +0.58 \\
    MultiHateClip & 74.40 & 75.01 & -0.61 \\
    \bottomrule
  \end{tabular}
\end{table}
As shown in Figure~\ref{fig:avg_rms_zcr}, RMS Energy reveals consistent differences between trimmed hate and trimmed non-hate segments across both datasets. Trimmed hate segments exhibit higher average energy levels—0.09-0.095 versus 0.08-0.085 in MultiHateClip-English, and 0.12-0.125 versus 0.09 in HateMM. This suggests hateful content is delivered with greater vocal intensity, though higher energy may also result from background audio or non-speech events rather than speech characteristics alone. In contrast, ZCR shows minimal sustained differences between segment types. Both trimmed hate and trimmed non-hate segments converge to similar spectral ranges (0.055-0.06 for MultiHateClip-English; 0.05-0.055 for HateMM), indicating that hate speech does not possess uniquely distinguishable acoustic signatures when isolated by temporal annotations.

The acoustic convergence between trimmed hate and trimmed non-hate segments extracted from hate videos reveals the contextual dependency of hate speech expression and the inadequacy of current temporal annotations. Despite attempts to isolate trimmed non-hate content, spectral similarity in ZCR indicates that the contextual dependency of hate speech extends beyond explicitly annotated boundaries. Acoustic contamination in trimmed non-hate segments reflects the fundamental challenge of defining precise temporal boundaries for hate content—hate expressions often involve gradual semantic transitions rather than discrete temporal shifts, while coarse-grained video-level annotations introduce systematic labelling noise.

\subsection{Visual Analysis}

To extend our multimodal investigation of label noise, we analyse visual content through object detection patterns across trimmed hate and trimmed non-hate segments. Following the methodology established by Das et al. [2], we employ the Ultralytics YOLOv8 \cite{yolov8_ultralytics} pre-trained model to detect objects in randomly sampled frames from each video segment, operating under the assumption that if an object appears in any frame of a given segment, it is present throughout that segment.
As shown in Table~\ref{tab:person_detection}, person detection rates demonstrate negligible differences between trimmed hate and trimmed non-hate segments, 0.58\% in HateMM and 0.61\% in MultiHateClip-English—indicating visual homogeneity across segment types.

This visual similarity demonstrates that hateful content exhibits continuity across visual contexts, with trimmed non-hate segments occurring within the same presentational environments as explicitly hateful material. The inability to distinguish segment types through visual object presence indicates that hate speech expression extends beyond discrete temporal boundaries, manifesting as contextually continuous phenomena rather than isolated incidents. These findings suggest that current temporal annotations operate at mismatched granularities. The visual evidence, combined with our other analyses, converges on a fundamental conclusion: existing annotation frameworks fail to accommodate the contextual dependencies and temporal continuity inherent in hate speech, resulting in systematic label noise that undermines detection model performance.

\begin{table*}[t]
\caption{Video classification performance (\%) on the HateMM and the English MultiHateClip subset datasets. MF1: Macro-F1; Acc: Accuracy; F1(H), P(H), R(H): F1, Precision, and Recall for the hate class. The best results are shown in \textbf{bold}.}
\centering
\begin{tabular}{lcccccccccccccc}
\toprule
\multirow{2}{*}{\textbf{Experiment Setting}} & \multicolumn{5}{c}{\textbf{HateMM}} & \multicolumn{5}{c}{\textbf{MultiHateClip (English)}} \\
& MF1 & Acc & F1(H) & P(H) & R(H) & MF1 & Acc & F1(H) & P(H) & R(H) \\
\midrule

\textit{Coarse Video-level Detection} & 79.30 & 79.77 & 76.46 & 73.90 & 79.49 & \text{64.37} & 68.38 & 52.60 & 54.21 & \text{51.30} \\

\textit{Noisy-to-Clean Generalization} & 76.91 & 77.23 & 79.10 & 77.48 & 81.35  & \text{57.63} & 66.33 & 38.89 & 52.74 & \text{32.43}  \\
\textit{Clean-to-Noisy Generalization} & 63.19 & 68.43 & 49.43 & 68.05 & 39.25  & \text{47.24} & 58.65 & 24.37 & 39.48 & \text{23.39}  \\
\textit{Clean Segment-Level Detection} & 98.64 & 98.66 & 98.80 & 99.23 & 99.37  & \text{97.31} & 97.55 & 96.51 & 98.28 & \text{94.82}  \\
\bottomrule
\end{tabular}
\label{tab:main_results}
\end{table*}

\begin{figure*}[t]
\centering

\begin{minipage}[t]{0.24\linewidth}
    \centering
    \includegraphics[width=\linewidth]{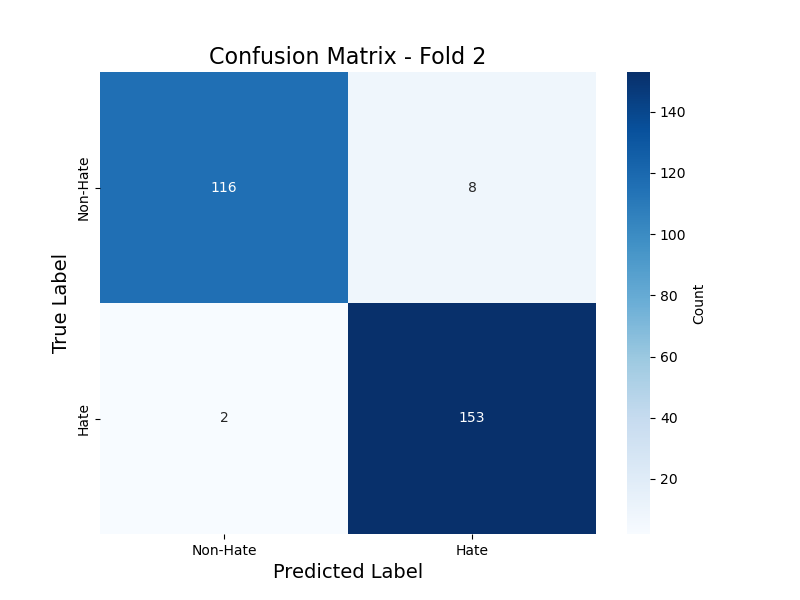}
    \textbf{(a)} HateMM – Clean Segment
\end{minipage}
\hfill
\begin{minipage}[t]{0.24\linewidth}
    \centering
    \includegraphics[width=\linewidth]{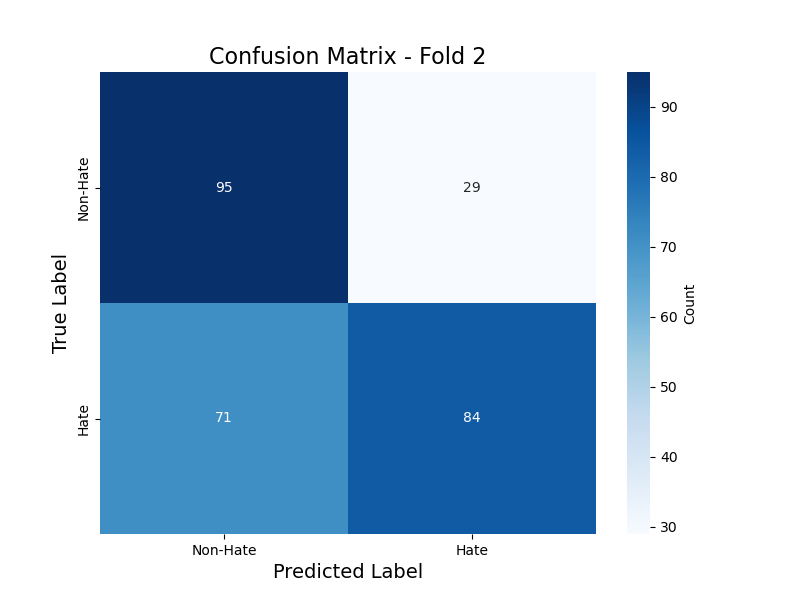}
    \textbf{(b)} HateMM – Noisy-to-Clean
\end{minipage}
\hfill
\begin{minipage}[t]{0.24\linewidth}
    \centering
    \includegraphics[width=\linewidth]{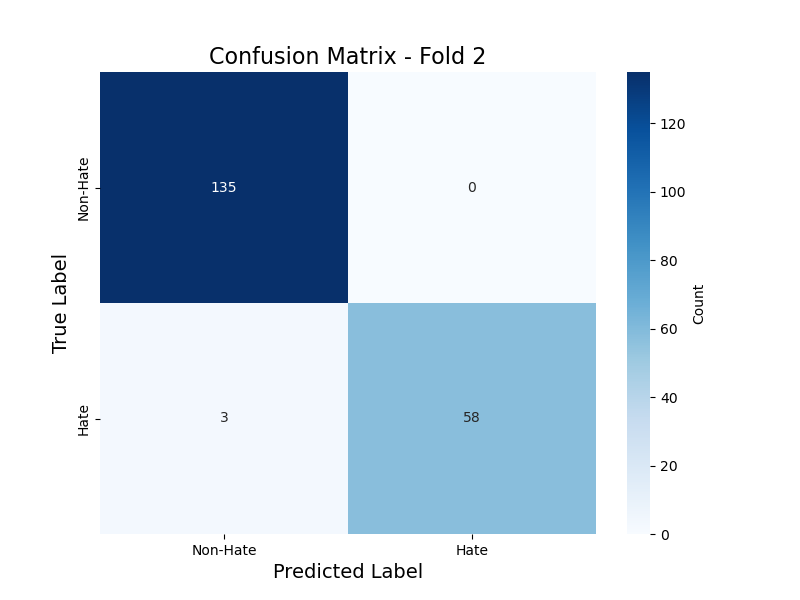}
    \textbf{(c)} MHC – Clean Segment
\end{minipage}
\hfill
\begin{minipage}[t]{0.24\linewidth}
    \centering
    \includegraphics[width=\linewidth]{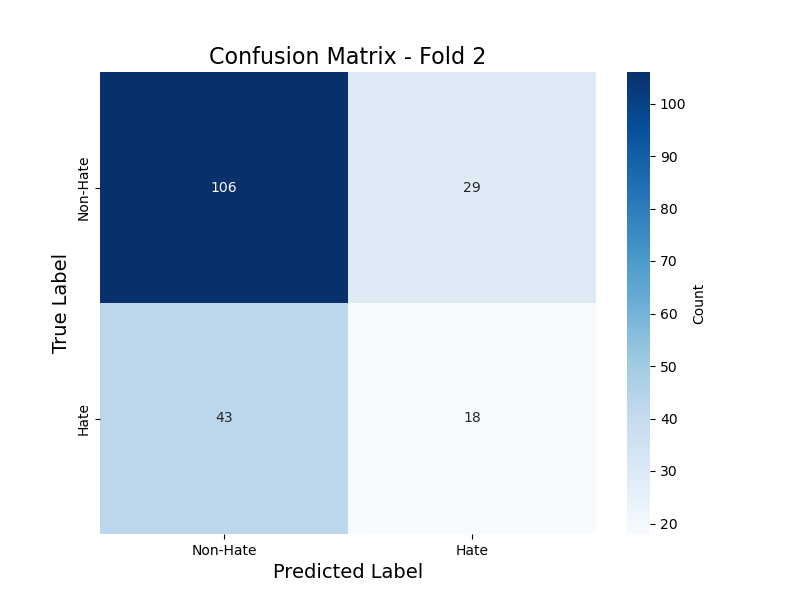}
    \textbf{(d)} MHC – Noisy-to-Clean
\end{minipage}

\caption{Confusion matrices comparing clean segment-level setting and noisy to clean setting. Models trained on noisy labels show degraded decision accuracy compared to clean-trained models, highlighting the impact of label noise on learning.}
\label{fig:confusion_matrices}
\end{figure*}

\subsection{Semantic Space Visualisation}
In order to explore the semantic boundaries between discourse types, we focus on text content, as it provides the most explicit semantic representation of hate speech, with most hateful intent typically conveyed through direct language expressions. We first extracted semantic features from transcript segments using Bidirectional Encoder Representations (BERT) \cite{bert} embeddings. These high-dimensional representations were then projected into a two-dimensional space using Uniform Manifold Approximation and Projection (UMAP) \cite{UMAP} to facilitate visualisation of semantic structure and overlap. Figure~\ref{fig:umap_semantic_space} presents semantic embedding space visualisations comparing different segment combinations across both datasets, offering critical insights into label noise and semantic drift.

The comparison between original video-level hate and non-hate labels (panels a and e) reveals substantial semantic overlap, with hate and non-hate content forming intermingled clusters. This overlap reflects the inherent label noise in video-level annotations, where videos labelled as hateful often contain significant non-hateful material. In contrast, comparisons between trimmed hate segments and non-hate videos (panels b and f) demonstrate clear semantic separation. Trimmed hate segments form distinct clusters apart from non-hate content, illustrating the benefit of temporal trimming in reducing noise and enhancing semantic discriminability. However, when comparing trimmed hate and trimmed non-hate segments drawn from the same hateful videos (panels c and g), considerable semantic overlap re-emerges. This indicates that within hateful videos, the boundaries between hateful and non-hateful content are not semantically discrete. Instead, the transition appears gradual, reflecting contextual dependencies and semantic drift across temporal sequences. Such continuity complicates the task of automated content detection.

Most notably, the three-way comparison (panels d and h) directly illustrates the problem of label noise. Trimmed non-hate segments do not cluster with non-hate videos but instead occupy an intermediate region overlapping with trimmed hate segments. This pattern provides empirical evidence that content labelled as non-hateful within hateful videos is semantically distinct from genuinely non-hateful content, underscoring how video-level labels introduce systematic contamination and obscure the fine-grained temporal dynamics of hate expression.

\section{Empirical Studies}
To systematically assess the impact of temporal label noise on multimodal hate detection, we design four complementary experimental configurations that isolate distinct aspects of the contamination phenomenon. This experimental framework is structured to evaluate (i) the degradation in detection performance caused by noisy temporal annotations, (ii) the performance improvements attainable through the use of temporally precise labels, and (iii) the influence of training data quality on model generalisation under both clean and noisy evaluation conditions.

As shown in Table \ref{tab:main_results}, the experimental design follows a controlled approach where we manipulate training and testing data composition whilst maintaining consistent model architectures and feature extraction pipelines. This enables direct attribution of performance differences to data quality rather than methodological variations. Our four experimental setups systematically vary the cleanliness of training and testing data:
\begin{itemize}
    \item Coarse Video-level Detection: Establishes baseline performance using conventional video-level annotations, representing current standard practice with inherent label noise. Full videos are used for training and testing.
    \item Noisy-to-Clean Generalisation: Evaluates whether models trained on noisy data can effectively identify trimmed hateful content, testing robustness to label contamination during training. Training uses full videos, testing uses trimmed hate segments and non-hate videos. 
    \item Clean-to-Noisy Generalisation: Assesses whether models trained on clean data can handle real-world noisy scenarios, testing practical deployment viability. Training uses trimmed hate segments and non-hate videos, testing uses full videos.
    \item Clean Segment-Level Detection: Demonstrates optimal performance potential using temporally-precise annotations, serving as an upper bound for detection accuracy. Training and testing use trimmed hate segments and non-hate videos.
\end{itemize}

\begin{figure*}[t]
    \centering
    \includegraphics[width=0.8\linewidth]{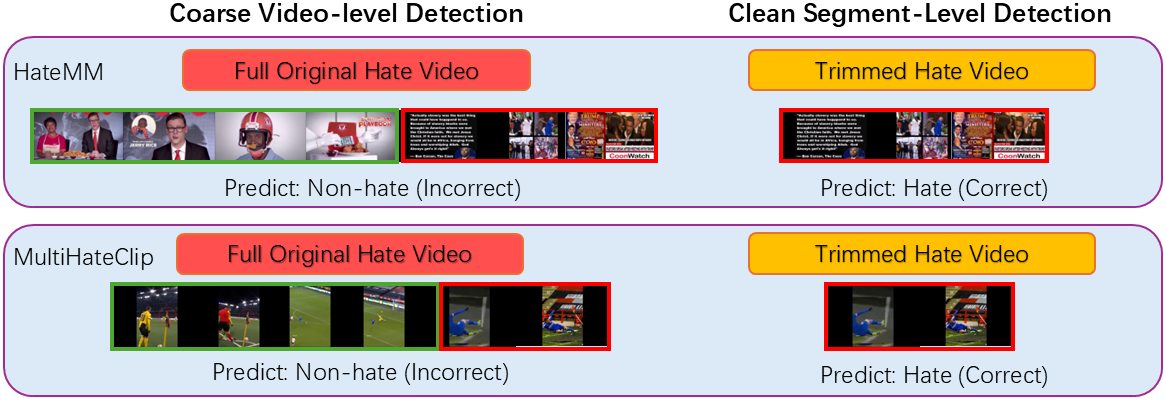}
    \caption{
    Case studies from the HateMM and MultiHateClip datasets demonstrate the model's prediction results in noisy and clean conditions.
    }
    \label{fig:casestudy}
\end{figure*}
\begin{figure*}[t]
    \centering
    \includegraphics[width=0.45\linewidth]{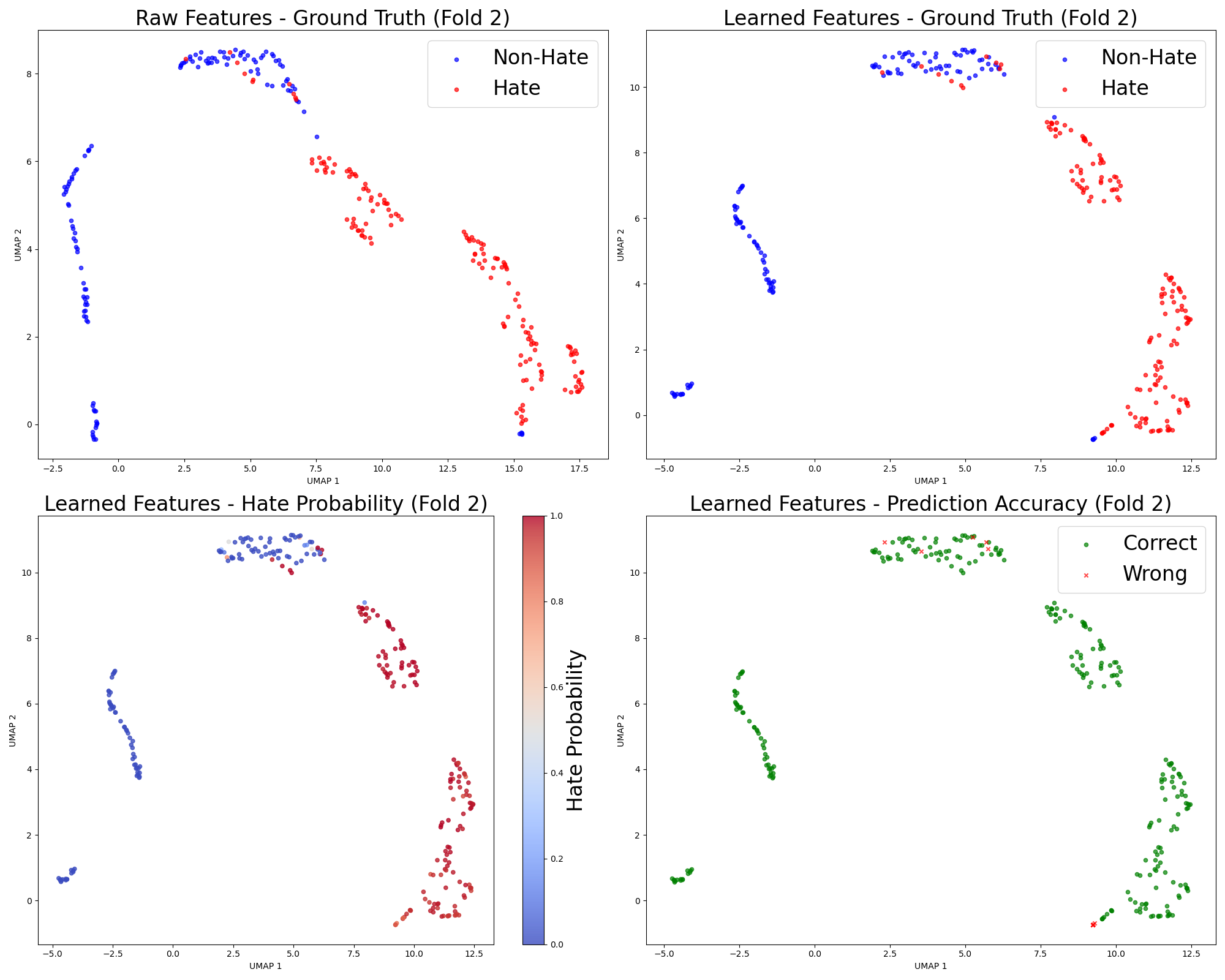}
    \includegraphics[width=0.45\linewidth]{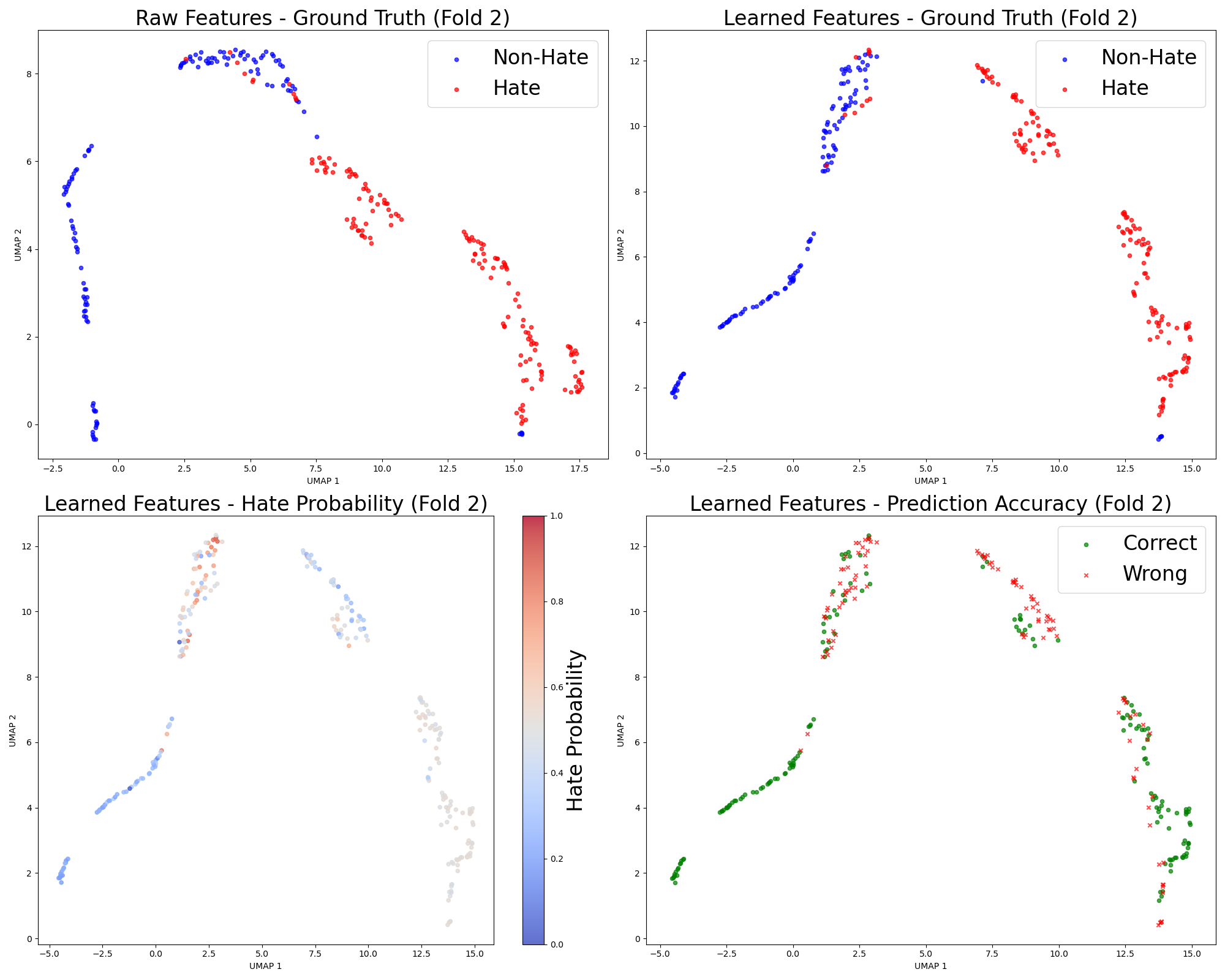}
    \caption{
    UMAP visualization of multimodal embeddings on clean segment-level setting and noisy to clean setting (HateMM).
    \textbf{Left:} Clean segment-level setting.
    \textbf{Right:} Noisy to clean setting.
    }
    \label{fig:umap_qualitative}
\end{figure*}
\subsection{Implementation Details}
We extract features from each trimmed hate and trimmed non-hate segment individually, treating them as independent samples. For non-hateful videos, we extract features from the full video without segmentation. Text, audio, and visual features are obtained using the same pipeline as in HateMM \cite{das2023hatemm} and MultiHateClip \cite{wang2024multihateclip}, which includes a BERT \cite{bert} model to obtain a 768-dimensional feature vector, extracting 40-dimensional Mel Frequency Cepstral Coefficients (MFCC) \cite{mfcc} from the audio, and processing uniformly sampled video frames with a Vision Transformer (Vit) \cite{vit} model \cite{dosovitskiy2021imageworth16x16words} to obtain 768-dimensional visual features.

We adopt a 70\%/10\%/20\% split for training, validation, and testing, and perform 5-fold cross-validation. We use standard evaluation metrics for this task: macro F1 score, accuracy, F1 score, precision, and recall. The best model is selected based on the macro F1 score in the validation set and evaluated on the test set. We adopt the baseline models from HateMM \cite{das2023hatemm} and MultiHateClip \cite{wang2024multihateclip} for all experiments. For both coarse video-level detection and clean segment-level detection, models are trained and evaluated for 20 epochs using a batch size of 16 and a learning rate of $1\text{e}^{-4}$. In the noisy-to-clean generalisation setting, the model is trained on full hate videos and evaluated on trimmed hate segments (HateMM: 40 epochs, batch size 128, learning rate $1\text{e}^{-5}$; MultiHateClip: 20 epochs, batch size 16, learning rate $1\text{e}^{-5}$). Conversely, in the clean-to-noisy generalisation setup, models are trained on trimmed hate segments and tested on full hate videos, using 20 training epochs, a batch size of 128, and a learning rate of $1\text{e}^{-5}$ for both datasets.

\subsection{Quantitative Results}

Table~\ref{tab:main_results} presents classification performance across all experimental configurations, demonstrating the substantial impact of temporal label noise on detection accuracy and model robustness. Clean segment-level detection significantly outperforms coarse video-level approaches, achieving macro F1 improvements of 19.34\% (HateMM) and 30.45\% (MultiHateClip-English). This performance gain directly quantifies the cost of label noise inherent in conventional video-level annotations.
The generalisation experiments reveal bidirectional robustness failures that illuminate the nature of label contamination. Models trained on noisy data struggle to identify trimmed hateful content, whilst models trained on clean data cannot effectively handle the temporal complexity present in contaminated videos. This asymmetric degradation demonstrates that label noise fundamentally alters model decision boundaries rather than introducing random classification errors.

Figure \ref{fig:confusion_matrices} presents confusion matrices comparing clean segment-level and noisy-to-clean configurations. Models trained on noisy data exhibit increased misclassification rates across both categories, with confusion matrices showing reduced diagonal dominance compared to clean-trained models. This pattern confirms that label noise undermines classification confidence and decision-making reliability, affecting the model's ability to establish clear semantic boundaries between hate and non-hate content.

\subsection{Qualitative Analysis}
To illustrate the impact of label noise, Figure~\ref{fig:casestudy} presents representative cases comparing coarse video-level and clean segment-level training. In both HateMM and MultiHateClip, models trained with coarse labels misclassify hateful content as non-hate, whereas clean-trained models correctly identify it. This highlights how label noise degrades the model's ability to detect hate speech. To further investigate this, Figure~\ref{fig:umap_qualitative} visualises UMAP embeddings from both settings on the same clean test set. While trimmed hate and non-hate segments remain semantically separable under both models, only clean-trained models yield confident predictions. In contrast, noisy-trained models show widespread uncertainty and frequent misclassifications (red crosses), despite similar embedding structures. This confirms that label noise mainly disrupts decision boundaries rather than representation learning, ultimately impairing reliable hate detection.

\section{Conclusion}

This study systematically investigates temporal label noise in multimodal hateful video classification. Our analysis reveals that 58.64\% of hate videos in HateMM and 35.16\% in MultiHateClip-English contain non-hateful material. Clean segment-level detection achieves macro F1 improvements of 19.34\% and 30.45\% over noisy video-level approaches. Data exploratory analysis demonstrates that segments from the same video exhibit significant semantic, acoustic, and visual overlap, providing empirical evidence for systematic rather than random label contamination. This label noise reflects the inadequacy of video-level annotations for capturing hate speech's localised nature. Hate content demonstrates context-dependent characteristics with gradual transitions from neutral to hateful expression, creating semantic drift that current temporal boundaries cannot accommodate.

These findings highlight the urgent need for temporally-aware detection models and annotation strategies that account for hate speech's contextual dependencies and temporal continuity.

\begin{acks}
This work was supported by the Alan Turing Institute and DSO National Laboratories Framework Grant Funding. We are grateful to Rishav Bhattacharyya  for his valuable insights and suggestions on this work.
\end{acks}

\bibliographystyle{ACM-Reference-Format}
\balance
\bibliography{sample-base}


\begin{thebibliography}{23}


\ifx \showCODEN    \undefined \def \showCODEN     #1{\unskip}     \fi
\ifx \showISBNx    \undefined \def \showISBNx     #1{\unskip}     \fi
\ifx \showISBNxiii \undefined \def \showISBNxiii  #1{\unskip}     \fi
\ifx \showISSN     \undefined \def \showISSN      #1{\unskip}     \fi
\ifx \showLCCN     \undefined \def \showLCCN      #1{\unskip}     \fi
\ifx \shownote     \undefined \def \shownote      #1{#1}          \fi
\ifx \showarticletitle \undefined \def \showarticletitle #1{#1}   \fi
\ifx \showURL      \undefined \def \showURL       {\relax}        \fi
\providecommand\bibfield[2]{#2}
\providecommand\bibinfo[2]{#2}
\providecommand\natexlab[1]{#1}
\providecommand\showeprint[2][]{arXiv:#2}

\bibitem[Andrews and Hofmann(2003)]%
        {MIL_label_noise_1}
\bibfield{author}{\bibinfo{person}{Stuart Andrews} {and} \bibinfo{person}{Thomas Hofmann}.} \bibinfo{year}{2003}\natexlab{}.
\newblock \showarticletitle{Multiple instance learning via disjunctive programming boosting}. In \bibinfo{booktitle}{\emph{Proceedings of the 17th International Conference on Neural Information Processing Systems}} (Whistler, British Columbia, Canada) \emph{(\bibinfo{series}{NIPS'03})}. \bibinfo{publisher}{MIT Press}, \bibinfo{address}{Cambridge, MA, USA}, \bibinfo{pages}{65–72}.
\newblock


\bibitem[Chen et~al\mbox{.}(2012)]%
        {Chen_2012}
\bibfield{author}{\bibinfo{person}{Ying Chen}, \bibinfo{person}{Yilu Zhou}, \bibinfo{person}{Sencun Zhu}, {and} \bibinfo{person}{Heng Xu}.} \bibinfo{year}{2012}\natexlab{}.
\newblock \showarticletitle{Detecting Offensive Language in Social Media to Protect Adolescent Online Safety}. In \bibinfo{booktitle}{\emph{2012 International Conference on Privacy, Security, Risk and Trust and 2012 International Confernece on Social Computing}}. \bibinfo{pages}{71--80}.
\newblock
\href{https://doi.org/10.1109/SocialCom-PASSAT.2012.55}{doi:\nolinkurl{10.1109/SocialCom-PASSAT.2012.55}}


\bibitem[Corazza et~al\mbox{.}(2020)]%
        {RNN_2020}
\bibfield{author}{\bibinfo{person}{Michele Corazza}, \bibinfo{person}{Stefano Menini}, \bibinfo{person}{Elena Cabrio}, \bibinfo{person}{Sara Tonelli}, {and} \bibinfo{person}{Serena Villata}.} \bibinfo{year}{2020}\natexlab{}.
\newblock \showarticletitle{A Multilingual Evaluation for Online Hate Speech Detection}.
\newblock \bibinfo{journal}{\emph{ACM Trans. Internet Technol.}} \bibinfo{volume}{20}, \bibinfo{number}{2}, Article \bibinfo{articleno}{10} (\bibinfo{date}{March} \bibinfo{year}{2020}), \bibinfo{numpages}{22}~pages.
\newblock
\showISSN{1533-5399}
\href{https://doi.org/10.1145/3377323}{doi:\nolinkurl{10.1145/3377323}}


\bibitem[Das et~al\mbox{.}(2023)]%
        {das2023hatemm}
\bibfield{author}{\bibinfo{person}{Mithun Das}, \bibinfo{person}{Rohit Raj}, \bibinfo{person}{Punyajoy Saha}, \bibinfo{person}{Binny Mathew}, \bibinfo{person}{Manish Gupta}, {and} \bibinfo{person}{Animesh Mukherjee}.} \bibinfo{year}{2023}\natexlab{}.
\newblock \showarticletitle{HateMM: A Multi-Modal Dataset for Hate Video Classification}. In \bibinfo{booktitle}{\emph{Proceedings of the International AAAI Conference on Web and Social Media}}, Vol.~\bibinfo{volume}{17}. \bibinfo{pages}{1014--1023}.
\newblock


\bibitem[Davidson et~al\mbox{.}(2017)]%
        {Davidson_2017}
\bibfield{author}{\bibinfo{person}{Thomas Davidson}, \bibinfo{person}{Dana Warmsley}, \bibinfo{person}{Michael Macy}, {and} \bibinfo{person}{Ingmar Weber}.} \bibinfo{year}{2017}\natexlab{}.
\newblock \showarticletitle{Automated Hate Speech Detection and the Problem of Offensive Language}.
\newblock \bibinfo{journal}{\emph{Proceedings of the International AAAI Conference on Web and Social Media}}  \bibinfo{volume}{11} (\bibinfo{date}{03} \bibinfo{year}{2017}).
\newblock
\href{https://doi.org/10.1609/icwsm.v11i1.14955}{doi:\nolinkurl{10.1609/icwsm.v11i1.14955}}


\bibitem[Devlin et~al\mbox{.}(2019)]%
        {bert}
\bibfield{author}{\bibinfo{person}{Jacob Devlin}, \bibinfo{person}{Ming-Wei Chang}, \bibinfo{person}{Kenton Lee}, {and} \bibinfo{person}{Kristina Toutanova}.} \bibinfo{year}{2019}\natexlab{}.
\newblock \showarticletitle{{BERT}: Pre-training of Deep Bidirectional Transformers for Language Understanding}. In \bibinfo{booktitle}{\emph{Proceedings of the 2019 Conference of the North {A}merican Chapter of the Association for Computational Linguistics: Human Language Technologies, Volume 1 (Long and Short Papers)}}, \bibfield{editor}{\bibinfo{person}{Jill Burstein}, \bibinfo{person}{Christy Doran}, {and} \bibinfo{person}{Thamar Solorio}} (Eds.). \bibinfo{publisher}{Association for Computational Linguistics}, \bibinfo{address}{Minneapolis, Minnesota}, \bibinfo{pages}{4171--4186}.
\newblock
\href{https://doi.org/10.18653/v1/N19-1423}{doi:\nolinkurl{10.18653/v1/N19-1423}}


\bibitem[Dosovitskiy et~al\mbox{.}(2020)]%
        {vit}
\bibfield{author}{\bibinfo{person}{Alexey Dosovitskiy}, \bibinfo{person}{Lucas Beyer}, \bibinfo{person}{Alexander Kolesnikov}, \bibinfo{person}{Dirk Weissenborn}, \bibinfo{person}{Xiaohua Zhai}, \bibinfo{person}{Thomas Unterthiner}, \bibinfo{person}{Mostafa Dehghani}, \bibinfo{person}{Matthias Minderer}, \bibinfo{person}{Georg Heigold}, \bibinfo{person}{Sylvain Gelly}, {et~al\mbox{.}}} \bibinfo{year}{2020}\natexlab{}.
\newblock \showarticletitle{An image is worth 16x16 words: Transformers for image recognition at scale}.
\newblock \bibinfo{journal}{\emph{arXiv preprint arXiv:2010.11929}} (\bibinfo{year}{2020}).
\newblock


\bibitem[Dosovitskiy et~al\mbox{.}(2021)]%
        {dosovitskiy2021imageworth16x16words}
\bibfield{author}{\bibinfo{person}{Alexey Dosovitskiy}, \bibinfo{person}{Lucas Beyer}, \bibinfo{person}{Alexander Kolesnikov}, \bibinfo{person}{Dirk Weissenborn}, \bibinfo{person}{Xiaohua Zhai}, \bibinfo{person}{Thomas Unterthiner}, \bibinfo{person}{Mostafa Dehghani}, \bibinfo{person}{Matthias Minderer}, \bibinfo{person}{Georg Heigold}, \bibinfo{person}{Sylvain Gelly}, \bibinfo{person}{Jakob Uszkoreit}, {and} \bibinfo{person}{Neil Houlsby}.} \bibinfo{year}{2021}\natexlab{}.
\newblock \bibinfo{title}{An Image is Worth 16x16 Words: Transformers for Image Recognition at Scale}.
\newblock
\showeprint[arxiv]{2010.11929}~[cs.CV]
\urldef\tempurl%
\url{https://arxiv.org/abs/2010.11929}
\showURL{%
\tempurl}


\bibitem[Fast et~al\mbox{.}(2016)]%
        {Fast_2016}
\bibfield{author}{\bibinfo{person}{Ethan Fast}, \bibinfo{person}{Binbin Chen}, {and} \bibinfo{person}{Michael~S. Bernstein}.} \bibinfo{year}{2016}\natexlab{}.
\newblock \showarticletitle{Empath: Understanding Topic Signals in Large-Scale Text}. In \bibinfo{booktitle}{\emph{Proceedings of the 2016 CHI Conference on Human Factors in Computing Systems}} \emph{(\bibinfo{series}{CHI’16})}. \bibinfo{publisher}{ACM}, \bibinfo{pages}{4647–4657}.
\newblock
\href{https://doi.org/10.1145/2858036.2858535}{doi:\nolinkurl{10.1145/2858036.2858535}}


\bibitem[Gamb{\"a}ck and Sikdar(2017)]%
        {gamback-sikdar-2017-using}
\bibfield{author}{\bibinfo{person}{Bj{\"o}rn Gamb{\"a}ck} {and} \bibinfo{person}{Utpal~Kumar Sikdar}.} \bibinfo{year}{2017}\natexlab{}.
\newblock \showarticletitle{Using Convolutional Neural Networks to Classify Hate-Speech}. In \bibinfo{booktitle}{\emph{Proceedings of the First Workshop on Abusive Language Online}}, \bibfield{editor}{\bibinfo{person}{Zeerak Waseem}, \bibinfo{person}{Wendy Hui~Kyong Chung}, \bibinfo{person}{Dirk Hovy}, {and} \bibinfo{person}{Joel Tetreault}} (Eds.). \bibinfo{publisher}{Association for Computational Linguistics}, \bibinfo{address}{Vancouver, BC, Canada}, \bibinfo{pages}{85--90}.
\newblock
\href{https://doi.org/10.18653/v1/W17-3013}{doi:\nolinkurl{10.18653/v1/W17-3013}}


\bibitem[Ji et~al\mbox{.}(2023)]%
        {Prompt_based}
\bibfield{author}{\bibinfo{person}{Junhui Ji}, \bibinfo{person}{Wei Ren}, {and} \bibinfo{person}{Usman Naseem}.} \bibinfo{year}{2023}\natexlab{}.
\newblock \showarticletitle{Identifying Creative Harmful Memes via Prompt based Approach}. In \bibinfo{booktitle}{\emph{Proceedings of the ACM Web Conference 2023}} (Austin, TX, USA) \emph{(\bibinfo{series}{WWW '23})}. \bibinfo{publisher}{Association for Computing Machinery}, \bibinfo{address}{New York, NY, USA}, \bibinfo{pages}{3868–3872}.
\newblock
\showISBNx{9781450394161}
\href{https://doi.org/10.1145/3543507.3587427}{doi:\nolinkurl{10.1145/3543507.3587427}}


\bibitem[Jocher et~al\mbox{.}(2023)]%
        {yolov8_ultralytics}
\bibfield{author}{\bibinfo{person}{Glenn Jocher}, \bibinfo{person}{Ayush Chaurasia}, {and} \bibinfo{person}{Jing Qiu}.} \bibinfo{year}{2023}\natexlab{}.
\newblock \bibinfo{booktitle}{\emph{Ultralytics YOLOv8}}.
\newblock
\urldef\tempurl%
\url{https://github.com/ultralytics/ultralytics}
\showURL{%
\tempurl}


\bibitem[Koutlis et~al\mbox{.}(2023)]%
        {MemeFier}
\bibfield{author}{\bibinfo{person}{Christos Koutlis}, \bibinfo{person}{Manos Schinas}, {and} \bibinfo{person}{Symeon Papadopoulos}.} \bibinfo{year}{2023}\natexlab{}.
\newblock \bibinfo{title}{MemeFier: Dual-stage Modality Fusion for Image Meme Classification}.
\newblock
\showeprint[arxiv]{2304.02906}~[cs.CV]
\urldef\tempurl%
\url{https://arxiv.org/abs/2304.02906}
\showURL{%
\tempurl}


\bibitem[Lang et~al\mbox{.}(2025)]%
        {MoRE}
\bibfield{author}{\bibinfo{person}{Jian Lang}, \bibinfo{person}{Rongpei Hong}, \bibinfo{person}{Jin Xu}, \bibinfo{person}{Yili Li}, \bibinfo{person}{Xovee Xu}, {and} \bibinfo{person}{Fan Zhou}.} \bibinfo{year}{2025}\natexlab{}.
\newblock \showarticletitle{Biting Off More Than You Can Detect: Retrieval-Augmented Multimodal Experts for Short Video Hate Detection}. In \bibinfo{booktitle}{\emph{Proceedings of the ACM on Web Conference 2025}} (Sydney NSW, Australia) \emph{(\bibinfo{series}{WWW '25})}. \bibinfo{publisher}{Association for Computing Machinery}, \bibinfo{address}{New York, NY, USA}, \bibinfo{pages}{2763–2774}.
\newblock
\showISBNx{9798400712746}
\href{https://doi.org/10.1145/3696410.3714560}{doi:\nolinkurl{10.1145/3696410.3714560}}


\bibitem[Leung et~al\mbox{.}(2011)]%
        {leung_hangling_lable_noise}
\bibfield{author}{\bibinfo{person}{Thomas Leung}, \bibinfo{person}{Yang Song}, {and} \bibinfo{person}{John Zhang}.} \bibinfo{year}{2011}\natexlab{}.
\newblock \showarticletitle{Handling label noise in video classification via multiple instance learning}. In \bibinfo{booktitle}{\emph{2011 International Conference on Computer Vision}}. \bibinfo{pages}{2056--2063}.
\newblock
\href{https://doi.org/10.1109/ICCV.2011.6126479}{doi:\nolinkurl{10.1109/ICCV.2011.6126479}}


\bibitem[McInnes et~al\mbox{.}(2020)]%
        {UMAP}
\bibfield{author}{\bibinfo{person}{Leland McInnes}, \bibinfo{person}{John Healy}, {and} \bibinfo{person}{James Melville}.} \bibinfo{year}{2020}\natexlab{}.
\newblock \bibinfo{title}{UMAP: Uniform Manifold Approximation and Projection for Dimension Reduction}.
\newblock
\showeprint[arxiv]{1802.03426}~[stat.ML]
\urldef\tempurl%
\url{https://arxiv.org/abs/1802.03426}
\showURL{%
\tempurl}


\bibitem[Muda et~al\mbox{.}(2010)]%
        {mfcc}
\bibfield{author}{\bibinfo{person}{Lindasalwa Muda}, \bibinfo{person}{Mumtaj Begam}, {and} \bibinfo{person}{Irraivan Elamvazuthi}.} \bibinfo{year}{2010}\natexlab{}.
\newblock \showarticletitle{Voice recognition algorithms using mel frequency cepstral coefficient (MFCC) and dynamic time warping (DTW) techniques}.
\newblock \bibinfo{journal}{\emph{arXiv preprint arXiv:1003.4083}} (\bibinfo{year}{2010}).
\newblock


\bibitem[Pramanick et~al\mbox{.}(2021)]%
        {MOMENTA}
\bibfield{author}{\bibinfo{person}{Shraman Pramanick}, \bibinfo{person}{Shivam Sharma}, \bibinfo{person}{Dimitar Dimitrov}, \bibinfo{person}{Md~Shad Akhtar}, \bibinfo{person}{Preslav Nakov}, {and} \bibinfo{person}{Tanmoy Chakraborty}.} \bibinfo{year}{2021}\natexlab{}.
\newblock \bibinfo{title}{MOMENTA: A Multimodal Framework for Detecting Harmful Memes and Their Targets}.
\newblock
\showeprint[arxiv]{2109.05184}~[cs.MM]
\urldef\tempurl%
\url{https://arxiv.org/abs/2109.05184}
\showURL{%
\tempurl}


\bibitem[Viola et~al\mbox{.}(2005)]%
        {MILBoost}
\bibfield{author}{\bibinfo{person}{Paul Viola}, \bibinfo{person}{John~C. Platt}, {and} \bibinfo{person}{Cha Zhang}.} \bibinfo{year}{2005}\natexlab{}.
\newblock \showarticletitle{Multiple instance boosting for object detection}. In \bibinfo{booktitle}{\emph{Proceedings of the 19th International Conference on Neural Information Processing Systems}} (Vancouver, British Columbia, Canada) \emph{(\bibinfo{series}{NIPS'05})}. \bibinfo{publisher}{MIT Press}, \bibinfo{address}{Cambridge, MA, USA}, \bibinfo{pages}{1417–1424}.
\newblock


\bibitem[Wang et~al\mbox{.}(2025)]%
        {Wang2025}
\bibfield{author}{\bibinfo{person}{Han Wang}, \bibinfo{person}{Rui~Yang Tan}, {and} \bibinfo{person}{Roy Ka-Wei Lee}.} \bibinfo{year}{2025}\natexlab{}.
\newblock \showarticletitle{Cross-Modal Transfer from Memes to Videos: Addressing Data Scarcity in Hateful Video Detection}. In \bibinfo{booktitle}{\emph{Proceedings of the ACM on Web Conference 2025}} \emph{(\bibinfo{series}{WWW ’25})}. \bibinfo{publisher}{ACM}, \bibinfo{pages}{5255–5263}.
\newblock
\href{https://doi.org/10.1145/3696410.3714534}{doi:\nolinkurl{10.1145/3696410.3714534}}


\bibitem[Wang et~al\mbox{.}(2024)]%
        {wang2024multihateclip}
\bibfield{author}{\bibinfo{person}{Han Wang}, \bibinfo{person}{Tan~Rui Yang}, \bibinfo{person}{Usman Naseem}, {and} \bibinfo{person}{Roy Ka-Wei Lee}.} \bibinfo{year}{2024}\natexlab{}.
\newblock \showarticletitle{Multihateclip: A multilingual benchmark dataset for hateful video detection on youtube and bilibili}. In \bibinfo{booktitle}{\emph{Proceedings of the 32nd ACM International Conference on Multimedia}}. \bibinfo{pages}{7493--7502}.
\newblock


\bibitem[Zhang et~al\mbox{.}(2024)]%
        {CMFusion}
\bibfield{author}{\bibinfo{person}{Yinghui Zhang}, \bibinfo{person}{Tailin Chen}, \bibinfo{person}{Yuchen Zhang}, {and} \bibinfo{person}{Zeyu Fu}.} \bibinfo{year}{2024}\natexlab{}.
\newblock \showarticletitle{{ Enhanced Multimodal Hate Video Detection via Channel-wise and Modality-wise Fusion }}. In \bibinfo{booktitle}{\emph{2024 IEEE International Conference on Data Mining Workshops (ICDMW)}}. \bibinfo{publisher}{IEEE Computer Society}, \bibinfo{address}{Los Alamitos, CA, USA}, \bibinfo{pages}{183--190}.
\newblock
\href{https://doi.org/10.1109/ICDMW65004.2024.00030}{doi:\nolinkurl{10.1109/ICDMW65004.2024.00030}}


\bibitem[Zimmerman et~al\mbox{.}(2018)]%
        {zimmerman-etal-2018-improving}
\bibfield{author}{\bibinfo{person}{Steven Zimmerman}, \bibinfo{person}{Udo Kruschwitz}, {and} \bibinfo{person}{Chris Fox}.} \bibinfo{year}{2018}\natexlab{}.
\newblock \showarticletitle{Improving Hate Speech Detection with Deep Learning Ensembles}. In \bibinfo{booktitle}{\emph{Proceedings of the Eleventh International Conference on Language Resources and Evaluation ({LREC} 2018)}}, \bibfield{editor}{\bibinfo{person}{Nicoletta Calzolari}, \bibinfo{person}{Khalid Choukri}, \bibinfo{person}{Christopher Cieri}, \bibinfo{person}{Thierry Declerck}, \bibinfo{person}{Sara Goggi}, \bibinfo{person}{Koiti Hasida}, \bibinfo{person}{Hitoshi Isahara}, \bibinfo{person}{Bente Maegaard}, \bibinfo{person}{Joseph Mariani}, \bibinfo{person}{H{\'e}l{\`e}ne Mazo}, \bibinfo{person}{Asuncion Moreno}, \bibinfo{person}{Jan Odijk}, \bibinfo{person}{Stelios Piperidis}, {and} \bibinfo{person}{Takenobu Tokunaga}} (Eds.). \bibinfo{publisher}{European Language Resources Association (ELRA)}, \bibinfo{address}{Miyazaki, Japan}.
\newblock
\urldef\tempurl%
\url{https://aclanthology.org/L18-1404/}
\showURL{%
\tempurl}


\end{thebibliography}










\end{document}